\documentclass{article}

\usepackage{arxiv}

\usepackage[utf8]{inputenc} 
\usepackage[T1]{fontenc}    
\usepackage{hyperref}       
\usepackage{url}            
\usepackage{booktabs}       
\usepackage{amsfonts}       
\usepackage{nicefrac}       
\usepackage{microtype}      
\usepackage{lipsum}		
\usepackage{graphicx}
\usepackage{natbib}
\usepackage{doi}

\usepackage{amsmath,amssymb}
\usepackage{mathrsfs}
\usepackage{array}
\usepackage{textcomp}
\usepackage{stfloats}
\usepackage{url}
\usepackage{verbatim}
\usepackage{graphicx}
\usepackage{booktabs}
\usepackage{amsfonts}
\usepackage{makecell}
\usepackage{color}
\usepackage[figuresright]{rotating}
\usepackage{multirow}
\usepackage[normalem]{ulem}
\usepackage[caption=false,font=normalsize,labelfont=sf,textfont=sf]{subfig}
\newcommand{\tabincell}[2]{\begin{tabular}{@{}#1@{}}#2\end{tabular}}
\usepackage{caption}
\usepackage{subcaption}

\usepackage{tabularx}
\usepackage{multicol}
\usepackage{multirow}
\usepackage{threeparttable}
\usepackage{colortbl}
\usepackage{xcolor}
\usepackage{ulem}
\usepackage{gensymb} 

\title{Latent Watermark: Inject and Detect Watermarks in Latent Diffusion Space}

\author{
    Zheling Meng\\
    Center for Research on \\Intelligent Perception and Computing, \\
    State Key Laboratory of \\Multimodal Artificial Intelligence Systems, \\
    Institute of Automation, \\
    Chinese Academy of Sciences \\
    \texttt{zheling.meng@cripac.ia.ac.cn} \\
    \And
    Bo Peng \\
    Center for Research on \\Intelligent Perception and Computing, \\
    State Key Laboratory of \\Multimodal Artificial Intelligence Systems, \\
    Institute of Automation, \\
    Chinese Academy of Sciences \\
    \texttt{bo.peng@nlpr.ia.ac.cn} \\
    \And
    Jing Dong\thanks{Corresponding Author}  \\
    Center for Research on \\Intelligent Perception and Computing, \\
    State Key Laboratory of \\Multimodal Artificial Intelligence Systems, \\
    Institute of Automation, \\
    Chinese Academy of Sciences \\
    \texttt{jdong@nlpr.ia.ac.cn} \\
}

\renewcommand{\shorttitle}{Latent Watermark: Inject and Detect Watermarks in Latent Diffusion Space}

\hypersetup{
pdftitle={Latent Watermark: Inject and Detect Watermarks in Latent Diffusion Space},
pdfauthor={Zheling Meng, Bo Peng, Jing Dong},
pdfkeywords={Latent diffusion model, Watermark, Latent space},
}

\begin{document}
\maketitle
\vspace{1cm}
\begin{abstract}
  Watermarking is a tool for actively identifying and attributing the images generated by latent diffusion models. Existing methods face the dilemma of image quality and watermark robustness. Watermarks with superior image quality usually have inferior robustness against attacks such as blurring and JPEG compression, while watermarks with superior robustness usually significantly damage image quality. This dilemma stems from the traditional paradigm where watermarks are injected and detected in pixel space, relying on pixel perturbation for watermark detection and resilience against attacks. In this paper, we highlight that an effective solution to the problem is to both inject and detect watermarks in the latent diffusion space, and propose Latent Watermark with a progressive training strategy. It weakens the direct connection between quality and robustness and thus alleviates their contradiction. We conduct evaluations on two datasets and against 10 watermark attacks. Six metrics measure the image quality and watermark robustness. Results show that compared to the recently proposed methods such as StableSignature, StegaStamp, RoSteALS, LaWa, TreeRing, and DiffuseTrace, LW not only surpasses them in terms of robustness but also offers superior image quality. Our code will be available at https://github.com/RichardSunnyMeng/LatentWatermark.

\end{abstract}


\section{Introduction}
\label{sec:introduction}
Recently, latent diffusion models \cite{nichol2022glide, rombach2022high, gu2022vector} are developing rapidly and have made many important breakthroughs in high-fidelity image generation. Trained on a large-scale dataset, a latent diffusion model can generate images with high resolution and quality according to text descriptions \cite{rombach2022high}. By efficient and effective fine-tuning \cite{hu2021lora}, some methods can generate images on specific scenarios and subjects. Besides open-source models, many online platforms also provide similar services to the public. While bringing convenience to our daily lives and work, they inevitably bring information security risks to our society. For example, some individuals use diffusion models to fabricate fake news, creating social panic and disrupting social order \cite{barrett2023identifying}. And some people create and spread rumors to defame the reputation of others. Methods that can identify images generated by latent diffusion models and determine their sources are urgently needed.
 
There are two main routes for identifying generated images, i.e. passive identification and active identification \cite{raja2021active}. Passive identification methods learn and extract forgery traces left by generators in spatial or frequency domain \cite{wang2020cnn, qian2020thinking, chai2020makes}. However, they usually cannot accurately identify generators unseen in training sets \cite{corvi2023detection, lorenz2023detecting, ricker2022towards}. Besides, they cannot attribute users who access the services to generate images. In contrast, active identification injects an invisible watermark into generated images before releasing them \cite{fernandez2022watermarking, tancik2020stegastamp, xiong2023flexible, nguyen2023stable, bui2023rosteals, fernandez2023stable, zhao2023recipe, wen2023tree}. The watermarks, usually in the form of multi-bit messages, can help to identify generated images (identification task) and attribute their sources (attribution task) \cite{he2020high, zhong2020automated, fang2022end, qin2024print, li2024screen}. The watermarking methods for latent diffusion models are attracting more and more attention.

\begin{figure*}[tb]
  \centering
  \includegraphics[scale=0.25]{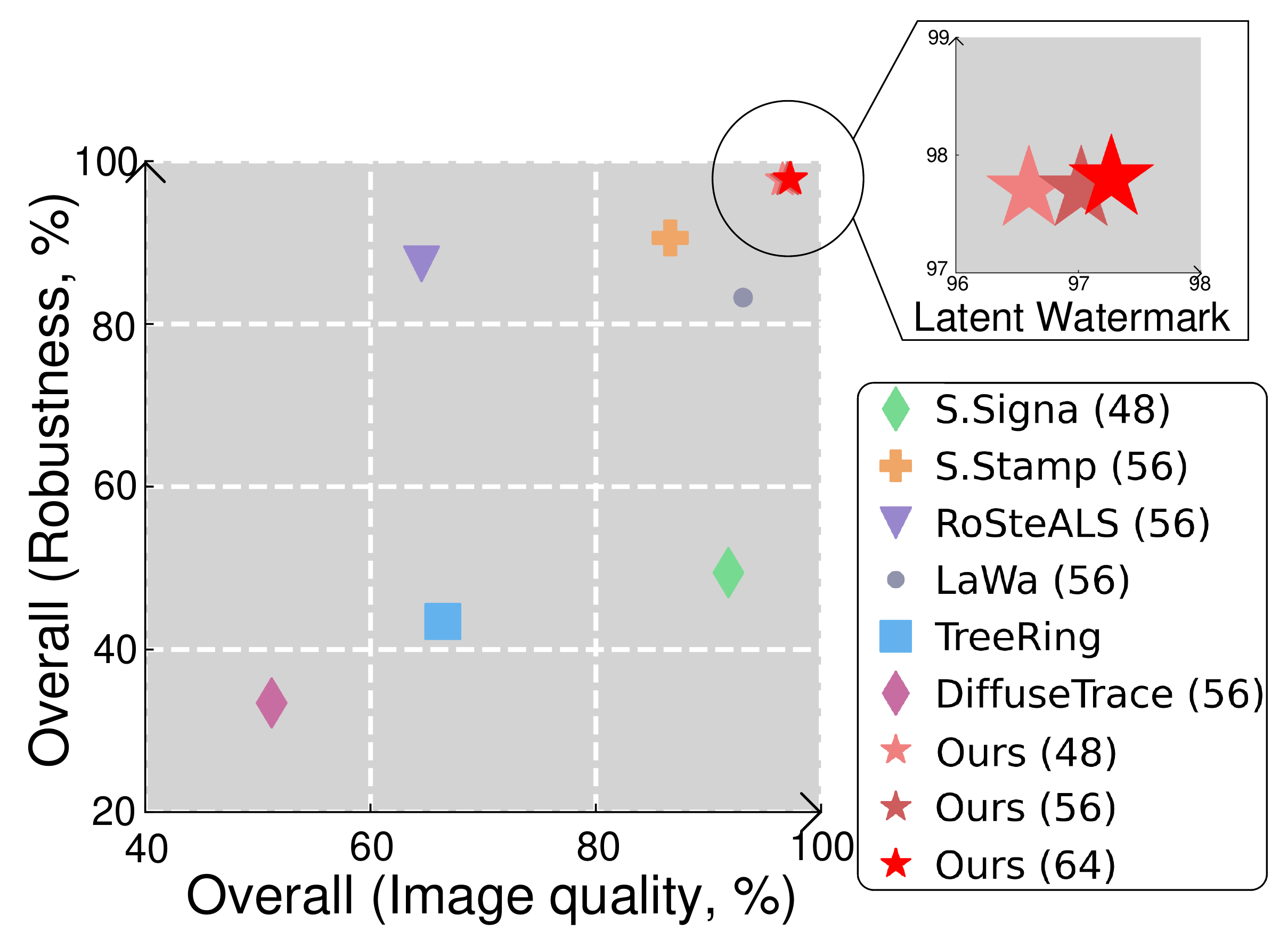}
  \caption{The overall performance of image quality and watermark robustness on the MS-COCO 2017 evaluation \cite{lin2014coco} for StableSignature (S.Signa., 48 bits) \cite{fernandez2023stable}, StegaStamp (S.Stamp, 56 bits) \cite{tancik2020stegastamp}, RoSteALS (56 bits) \cite{bui2023rosteals}, LaWa (48 bits) \cite{rezaei2024lawa}, TreeRing \cite{wen2023tree}, DiffuseTrace (56 bits) \cite{lei2024diffusetrace}, as well as our proposed Latent Watermark (48 bits, 56 bits, and 64 bits). Please see Sec.\ref{sec:setup} for more details about the overall performance calculation.}
  \label{fig:compare with other methods}
\end{figure*}

Existing watermarking methods for latent diffusion models can be classified into three categories, i.e. executing watermark algorithms before \cite{fernandez2023stable}, during \cite{nguyen2023stable, wen2023tree}, and after generating images \cite{tancik2020stegastamp, bui2023rosteals}. A good watermarking method should satisfy strong robustness and high image quality. Strong robustness helps resist attacks from intentional and unintentional damages. High image quality can alleviate the concerns among model developers regarding its potential negative impact on their generative models, thereby encouraging the widespread application of the method. The injection of watermarks should be done in an imperceptible manner and should not change or even limit the generation ability of vanilla models. However, existing methods often exhibit a trade-off between image quality and watermark robustness. We evaluate four previous methods that have been reported to have good performance, i.e. SteagStamp \cite{tancik2020stegastamp}, StableSignature \cite{fernandez2023stable}, RoSteALS \cite{bui2023rosteals}, and LaWa \cite{rezaei2024lawa}. We show their overall performance of image quality and watermark robustness in Fig.\ref{fig:compare with other methods}. Please refer to Sec.\ref{sec:setup} for more details about the results. It can be seen that SteagStamp and RoSteALS have better robustness but worse image quality, while StableSignature and LaWa has better quality but worse robustness. 

We highlight that the dilemma stems from \textbf{the watermarking paradigm} where watermarks are injected and detected in pixel space, which is used by most of existing methods \cite{fernandez2022watermarking, tancik2020stegastamp, xiong2023flexible, nguyen2023stable, bui2023rosteals, fernandez2023stable, zhao2023recipe}. The paradigm guides watermarking models to inject watermarks in a manner of pixel perturbation. When the magnitude of this perturbation is small, it is not enough to resist various attacks. When the magnitude is large, image quality is severely affected. 

In this paper, we propose \textbf{L}atent \textbf{W}atermark (LW) to watermark and detect images generated by latent diffusion models in latent space. Our motivation is that if watermarks are injected and detected in latent diffusion space, we can \textbf{avoid directly relying on pixel perturbation} for watermark detection and resilience against attacks. It weakens the correlation between image quality and watermark robustness with the latent encoder and decoder. 

\label{sec: intro}

However, our experiments reveal that injecting and detecting watermarks using LW with a high image quality and robustness is not easy. Our method inserts four modules for watermark injection and detection. The challenge mainly lies in training these modules. The analysis in Sec.\ref{sec: necessity} reveals that it is necessary to involve the latent encoder and decoder in the training to improve image quality and robustness. In the training phase, the latent images with watermarks are first decoded into the pixel images by the latent decoder, and then encoded into the latent images by the latent encoder for watermark detection. When employing random initialization for parameters, the initial parameters of these modules are distant from their globally optimal solutions. This discrepancy hinders the effective adaptation between the input/output of each module and the frozen latent encoder/decoder, leading the training process to converge towards local solutions. To address this challenge, in the manuscript, we propose a three-step progressive training strategy, which trains the watermark modules from local to global while freezing the weights of the vanilla model. The motivation stems from the idea of divide-and-conquer: first pre-train each module separately and then fine-tune them together. It provides initial points close to optimal solutions, serving as a more suitable starting point for formal training. It significantly reduces the difficulty of training and improves the training efficiency.

Some recent and concurrent studies also consider watermarking using the latent space. RoSteALS \cite{bui2023rosteals} and LaWa \cite{rezaei2024lawa} inject watermarks in the latent space but detect watermarks on the pixel images directly. It makes the watermarks still rely on pixel perturbation, leading to the difficulty in balancing robustness and image quality. TreeRing \cite{wen2023tree} and DiffuseTrace \cite{lei2024diffusetrace} write watermarks into the initial noises sampled in the latent space. They detect watermarks by restoring the initial noises through the forward diffusion process. However, the errors between the vanilla images and the attacked images accumulate during the diffusion process, resulting in low watermark robustness. Different from them, with the help of the proposed training strategy, our method injects and detects watermarks completely in the latent space and avoids the low robustness caused by the error accumulation.  Compared with these methods, it shows higher performance as presented in Fig.\ref{fig:compare with other methods}.

We evaluate the performance of our methods and the previous six methods using the captions collected from MS-COCO 2017 \cite{lin2014coco} and Flickr30k \cite{young2014image} datasets. Six metrics are measured in the experiments, including FID, SSIM, NIQE, and PIQE for evaluating image quality, and BitACC and TPR@0.01FPR for evaluating watermark robustness under 10 attacks. The attacks cover the destructive, constructive, and reconstructive attacks. The results illustrate that our method has better robustness and image quality compared to other methods, significantly alleviating the trade-off between them as shown in Fig.\ref{fig:compare with other methods}. We also study the effectiveness of each training step, the position for watermark injection, and the size of training data on the performance of LW. Extensive discussions are also conducted. 

In summary, the contributions of this paper are as follows.
\begin{itemize}
    \item Previous methods resort to perturbations in the pixel space for watermark injection, which have difficulty in resisting attacks while maintaining image quality. Several methods attempt to use the latent diffusion space, but they either continue to rely on pixel perturbations or struggle with inferior robustness. 
    \item We highlight injecting and detecting watermarks completely in latent diffusion space by inserting several tiny modules and propose Latent Watermark (LW). Further, we propose a progressive training strategy to address the challenge of effectively training them.
    \item We conduct evaluations using the captions from MS-COCO and Flickr30k datasets. 10 attacks from three categories are applied to the watermarked images. Based on six metrics, the results demonstrate that LW can inject more robust watermarks with higher image quality.
\end{itemize}

\section{Related Work}
\subsection{Latent Diffusion Model}
Diffusion probabilistic models \cite{sohl2015deep, ho2020denoising} are proposed to learn a data distribution $p(\widetilde{x})$ from a real distribution $q(x)$ by the Markov forward and backward diffusion process. Specifically, they train a noise predictor $\epsilon_\theta(\widetilde{x}_t, t)$ to generate an image $\widetilde{x}_0$ from a sampled Gaussian noise $\widetilde{x}_T$ by estimating the noises and performing denoising for $T$ steps. In order to speed up the generation process, Song et al. propose Denoising Diffusion Implicit Model (DDIM) \cite{song2020denoising} to reduce the standard 1000-step denoising process to fewer, usually 50 steps. To reduce computational resource requirements while retaining the quality, Latent Diffusion Model (LDM) \cite{rombach2022high} performs the diffusion process in a latent space and becomes a standard paradigm for image generation using diffusion models \cite{nichol2022glide, rombach2022high, gu2022vector}. LDM performs denoising in the latent space with lower resolution and then uses a latent decoder $Dec(\cdot)$ to generate human-understandable images with higher resolution:
\begin{equation}
    \begin{split}
        \widetilde{x} &= Dec(\widetilde{z}_0) \\
        \widetilde{z}_{t-1} &= \frac{1}{\sqrt{\alpha_t}}\left(\widetilde{z}_t - \frac{1-\alpha_t}{\sqrt{1-\overline{\alpha}_t}}\epsilon_\theta(\widetilde{z}_t, t) \right) \quad\quad (t = 1, 2, ..., T)
    \end{split}
\end{equation}
where $\alpha_t = 1-\beta_t$, $\overline{\alpha}_t = \prod^t_{i=1}\alpha_i$ and $\beta \in (0, 1)$ is a scheduled noise variance. For training the noise predictor $\epsilon_\theta(\widetilde{z}_t, t)$ in the latent space, LDM also includes a latent encoder $Enc(\cdot)$ to encode training images. In this paper, Stable Diffusion (SD) \cite{rombach2022high}, a classic model implementation for LDM, is used to introduce and evaluate our watermarking method. The shape of $\widetilde{z}_t$ and $\widetilde{x}$ is $(4, 64, 64)$ and $(3, 512, 512)$ respectively.

\subsection{Watermarks for Latent Diffusion Models}
Watermarking methods typically inject information as a series of bits within generated images in a subtly or imperceptibly manner. It allows for the determination of whether an image is generated and by which user through the detection of this injected information. For latent diffusion models, the methods can be classified into three categories according to the order of watermark algorithm execution and image generation.

Execute watermark algorithms before generation. The methods allow diffusion models to directly generate watermarked images without introducing any other modules. Zhao et al. \cite{zhao2023recipe} train unconditional or class-conditional models using watermarked training images, and generate watermarked images using a trigger prompt for text-conditional models. StableSignature \cite{fernandez2023stable} roots a watermark into model weights by training a message encoder-decoder and fine-tuning the diffusion model. Once training is completed, both methods cannot change the injected messages.

Execute watermark algorithms during generation. Stable Messenger \cite{nguyen2023stable} embeds an encoded bit message into latent diffusion space and decodes it from generated images. The work \cite{xiong2023flexible} carefully designs the fusion method for encoded messages and latent images. It also proposes a secure mechanism that can overcome watermark injection escape caused by simply commenting out the codes. As the most fundamental difference from our method, they both detect watermarks in pixel space via training another decoder. TreeRing \cite{wen2023tree} directly writes messages in the spectrum of sampled noises and detects the messages by Gaussian noising and spectral transformation. However, it fails to attribute users, limiting its application scenarios. The proposed LW belongs to this category.

Execute watermark algorithms post generation. Given a generated image, the methods generate a generator-independent watermark and inject it into the image. StegaStamp \cite{tancik2020stegastamp} follows this way and adopts many augmentation methods in the training stage to enhance its robustness. Inspired by self-supervised learning, Fernandez et al. \cite{fernandez2022watermarking} propose SSL-Watermarking to optimize an invisible watermark image-by-image and detect it by estimating vector cosine angles. RoSteALS \cite{bui2023rosteals} encodes input images into a latent space by a well trained autoencoder and then injects encoded messages. Same as \cite{nguyen2023stable} and \cite{xiong2023flexible}, the methods also detect watermarks in pixel space via training another decoder.

\subsection{Watermark Attacks}
\label{sec: attack methods}
The purpose of watermark attack experiments is to evaluate the robustness of watermarking methods when faced with malicious or unintentional image corruptions in practice. The common attacks can be classified into three categories, i.e. destructive attacks, constructive attacks and reconstructive attacks \cite{zhao2023generative}. Destructive attacks include brightness distortion, contrast distortion, JPEG compression and Gaussian noising. Constructive attacks mainly include some denoising algorithms using Gaussian kernels or Block-Matching and 3D filtering (BM3D) \cite{dabov2007image}. Recently, Zhao et al. propose reconstructive attacks \cite{zhao2023generative} to erase watermarks. They use an image reconstruction model, such as a diffusion model or a variational autoencoder, to encode semantic features of an image and regenerate it. The work highlights that a simple reconstruction model can erase watermarks injected by most existing methods.
\section{Methods}

\begin{figure}[t]
  \centering
  \includegraphics[scale=0.3]{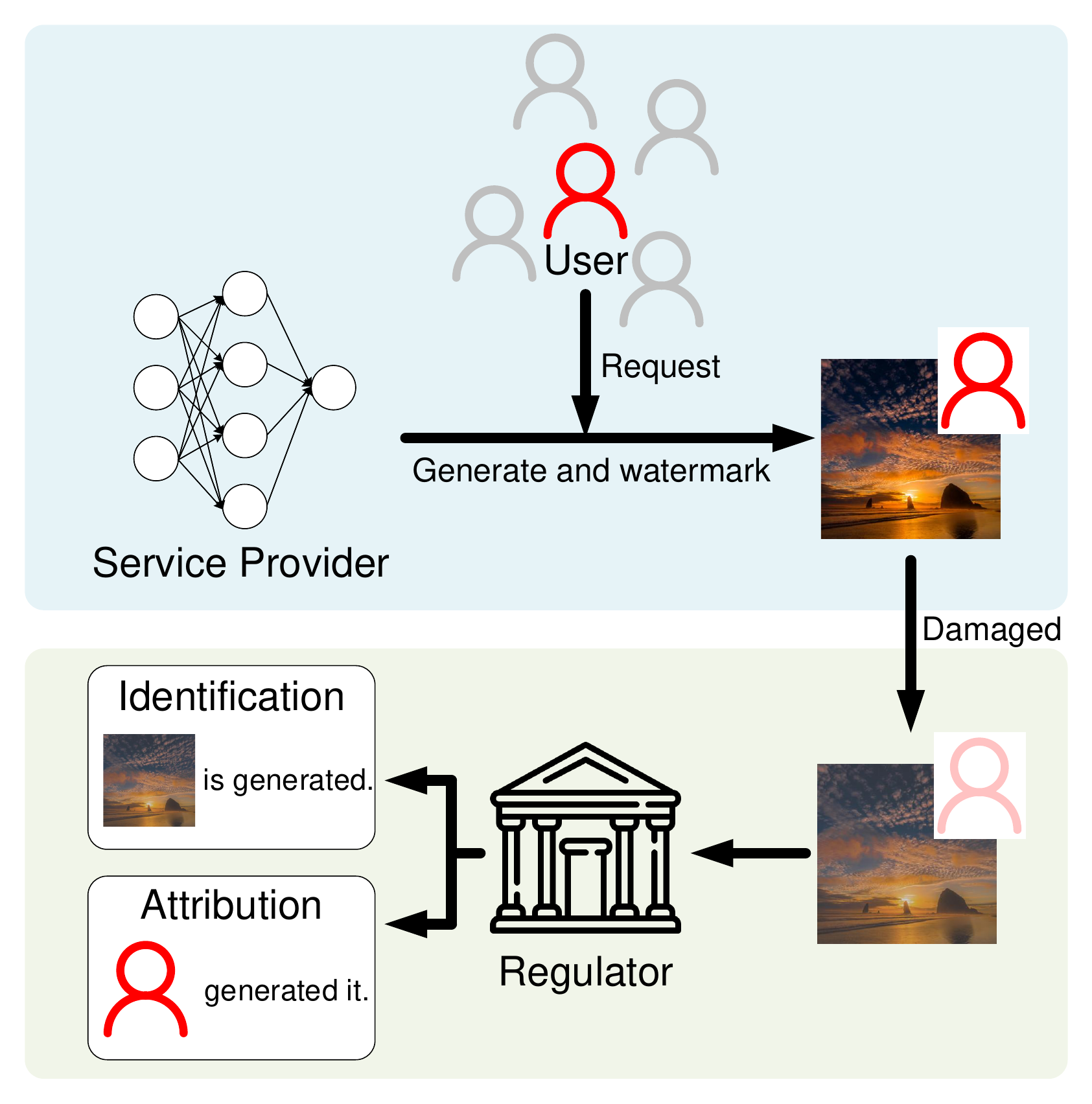}
  \caption{The threat model. The service provider generates an image according to the request of the user and watermarks it. The regulator detects the watermark of the damaged image for the identification and attribution task.}
  \label{fig: threat}
\end{figure}

\subsection{Threat Model}
Fig.\ref{fig: threat} visualizes our threat model. There are three agents, i.e. a service provider, a user, and a regulator. Owning a latent diffusion model, the service provider provides image generation services to the public through an API. The generated images are watermarked before being released. Given a text description, the user calls the diffusion model through the API to obtain generated images. The regulator, from the service provider or government departments, completes the following two tasks by detecting watermarks:
\begin{itemize}
    \item Identification: Determine whether an image is generated by the service provider.
    \item Attribution: Determine which user generates the image through the API.
\end{itemize}
The difference between the two is that the latter requires an exact match of bits. Due to unintentional and intentional damages, there are differences in image quality between generated and detected images but the contents of detected images can still be accurately understood by humans. We formally summarize the threat model as follows. We denote $\widetilde{x}$ as a generated image, distinguished from a real image $x$.
\begin{itemize}
    \item Service Provider: She owns a latent diffusion model $\epsilon_\theta$ and provides an API to the public to generate images watermarked by an algorithm $\mathcal{T}$. $\mathcal{T}$ injects identity information in bit form into generated images while maintaining image quality so that it cannot be perceived by users.
    \item User: She obtains an image $\widetilde{x}$ conditioned on a text description $c$ through the API. $\widetilde{x}$ can be changed into $\widetilde{x}^\prime$ by common attacks as mentioned in Sec.\ref{sec: attack methods}. And humans have a consistent understanding of $\widetilde{x}$ and $\widetilde{x}^\prime$.
    \item Regulator: Given $\mathcal{T}$, the regulator tries to identify and attribute $\widetilde{x}^\prime$.
\end{itemize}

\begin{figure}[tb]
  \centering
  \includegraphics[scale=0.15]{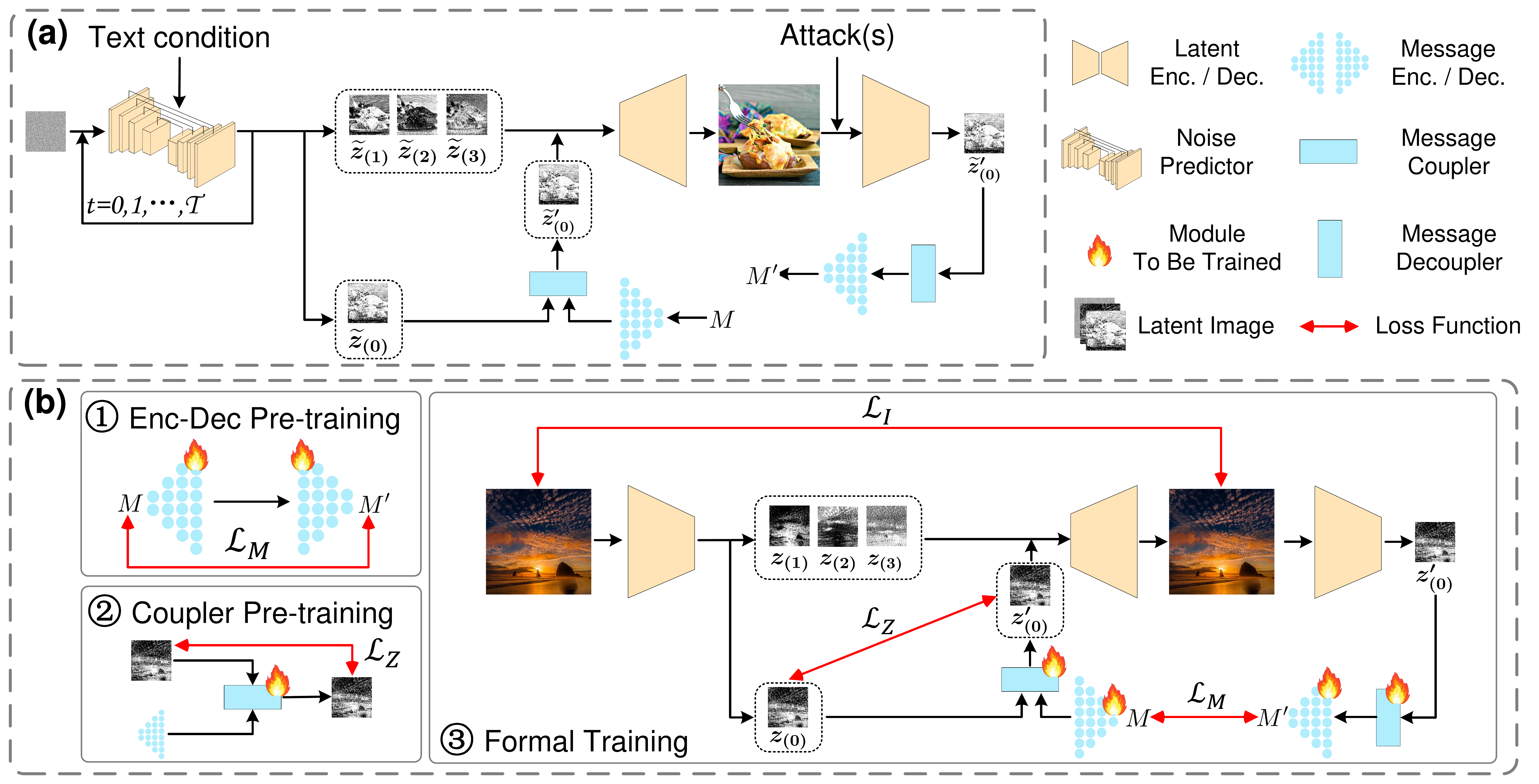}
  \caption{The proposed methods. (a) The structure of LW. (b) The three-step progressive training strategy. $M$: $n$-bit messages. $z_{(l)}$: $l$-th channel of latent image $z$.}
  \label{fig:methods}
\end{figure}

\subsection{Latent Watermark}
Fig.\ref{fig:methods} (a) shows the structure of LW. LW is composed of a message encoder $Enc_M(\cdot)$, a message coupler $C(\cdot, \cdot)$, a message decoupler $DC(\cdot)$ and a message decoder $Dec_M(\cdot)$. Besides, a latent diffusion model, including a latent encoder $Enc(\cdot)$, a noise predictor $\epsilon_\theta(\widetilde{z}_t, t)$ and a latent decoder $Dec(\cdot)$, is also needed. 

In the stage of injecting watermarks, given a $n$-bit message, $Enc_M(\cdot)$ first encodes it into the latent space. Then, $C(\cdot, \cdot)$ fuses the latent image $\widetilde{z}$ from $\epsilon_\theta(\widetilde{z}_t, t)$ with the encoded message to obtain $\widetilde{z}^\prime$. Next, $\widetilde{z}^\prime$ is decoded by $Dec(\cdot)$ to generate an image $\widetilde{x}$. The injecting process can be expressed as Eq.\ref{eq:inject watermark}:
\begin{equation}
\label{eq:inject watermark}
    \begin{split}
        \widetilde{x} = Dec\left( C(\widetilde{z}, Enc_M\left( M \right)) \right)
    \end{split}
\end{equation}
where $M\in \{0,1\}^n$ is a $n$-bit message. In the experiments, we find that choosing different channels to fuse messages will bring different performance. According to the results in Tab.\ref{tab: channels}, we design $C(\cdot, \cdot)$ to fuse the first channel of $\widetilde{z}$ with the encoded messages, as shown in Fig.\ref{fig:methods} (a) and Eq.\ref{eq: channel}:

\begin{equation}
\label{eq: channel}
    \begin{split}
        \widetilde{z}^\prime &= C(\widetilde{z}, Enc_M\left( M \right)) \\
        &= C(\widetilde{z}_{(0)}, Enc_M\left( M \right)) \oplus \widetilde{z}_{(1)} \oplus \widetilde{z}_{(2)} \oplus \widetilde{z}_{(3)}
    \end{split}
\end{equation}
where $\oplus$ is concatenation along the channel dimension and $\widetilde{z}_{(l)}$ is $l$-th channel of $\widetilde{z}$ ($l=\{0, 1, 2, 3\}$). $Enc_M\left( M \right)$ has the same shape with $\widetilde{z}_{(l)}$.

In the stage of detecting watermarks, $Enc(\cdot)$ is used to encode an image $\widetilde{x}^\prime$ into the latent space. Then $DC(\cdot)$ decouples the encoded message from the first channel of the latent image. Finally, $Dec_M(\cdot)$ outputs the decoding result in bit form. The detecting process can be expressed as Eq.\ref{eq:detect watermark}:
\begin{equation}
\label{eq:detect watermark}
    \begin{split}
        M^\prime &= Dec_M\left( DC(\widetilde{z}^\prime_{(0)}) \right) \\
        \widetilde{z}^\prime &= Enc\left( \widetilde{x}^\prime \right)
    \end{split}
\end{equation}
where $M^\prime \in \{0,1\}^n$ is the decoded message.

\subsection{Three-Step Progressive Training Strategy}
As pointed out in Sec.\ref{sec:introduction}, the strategy to train LW is the key to obtain working performance. Here, we propose a progressive training strategy. Experiments in Sec.\ref{sec:ablation} show that without it, watermark injection and image quality maintenance cannot be reached at the same time. As shown in Fig.\ref{fig:methods} (b), the training strategy includes three steps, i.e. pre-training the message encoder and decoder, pre-training the message coupler and training the model formally. For the purpose of efficiency, $z$ from training images encoded by $Enc(\cdot)$, rather than $\widetilde{z}$ from $\epsilon_\theta(\widetilde{z}_t, t)$, is used in the training stage.

\textbf{Step 1: Pre-train the message encoder and decoder.} In this step, the encoded messages are input into the message decoder directly to train them in the self-supervision paradigm. The loss function in this step is $\mathcal{L}_{1}$ in Eq.\ref{eq:step1}:
\begin{equation}
\label{eq:step1}
\begin{split}
    \mathcal{L}_{1} &= \frac{1}{B \cdot n}\sum_{i=0}^{B-1} \sum_{k=0}^{n-1} \left( 
M_i^k - {M_i^{k}}^\prime \right)^2 \\
\end{split}
\end{equation}
where $M_i \in \{0,1\}^n$ is the $i$-th randomly generated message, $M_i^{k}$ is $k$-th bit of $M_i$ and $B$ is the batch size. When the exponential moving average of Bit Accuracy is higher than the threshold $\tau_1$, the step ends.

\textbf{Step 2: Pre-train the message coupler.} LW is expected to inject watermarks with minimizing any impact on image quality. Therefore, optimizing the parameters of the modules should be started from the initial state, i.e., outputting the same images as the vanilla model regardless of encoded messages. In this step, the coupler will be pre-trained as an identity mapping of latent images. The loss function $\mathcal{L}_{2}$ can be expressed as Eq.\ref{eq:step2}:
\begin{equation}
\label{eq:step2}
\begin{split}
    \mathcal{L}_{2} &= \frac{1}{|\mathcal{D}|}\frac{1}{WH}\sum_{i\in\mathcal{D}} ||z_{i(0)} - z_{i(0)}^\prime||_2^2
\end{split}
\end{equation}
where $\mathcal{D}$ is the training batch with the batch size $|\mathcal{D}|$, $W$ and $H$ are the width and height of $z_i$, and $z^\prime_i$ is obtained using the same method outlined in Eq.\ref{eq: channel}. When $\mathcal{L}_{2}$ is lower than the threshold $\tau_2$, the step ends.

\textbf{Step 3: Train the model formally.}
In this step, three loss functions, i.e. $\mathcal{L}_z$, $\mathcal{L}_I$ and $\mathcal{L}_M$, are used to train all the watermark-related modules. As shown in Eq.\ref{eq:step4-latent}, $\mathcal{L}_z$ has the same form as Eq.\ref{eq:step2} and measures the L2 distance of latent images before and after message coupling:
\begin{equation}
\label{eq:step4-latent}
    \mathcal{L}_z = \frac{1}{|\mathcal{D}|}\frac{1}{WH}\sum_{i\in \mathcal{D}} ||z_{i(0)} - z_{i(0)}^\prime||^2_2.
\end{equation}
And $\mathcal{L}_I$ measures the visual similarity between images before and after injecting watermarks as shown in Eq.\ref{eq:step4-lpips}:
\begin{equation}
\label{eq:step4-lpips}
    \mathcal{L}_I = \frac{1}{|\mathcal{D}|}\sum_{i\in \mathcal{D}} LPIPS(x_i, Dec(z^\prime_i))
\end{equation}
where $LPIPS(\cdot, \cdot)$ is a commonly used method for visual similarity evaluation \cite{zhang2018unreasonable}. Inspired by \cite{nguyen2023stable}, we supervise the message encoding and decoding by Eq.\ref{eq:step4-bit loss}:
\begin{equation}
\label{eq:step4-bit loss}
\resizebox{0.68\hsize}{!}{$
    \mathcal{L}_M = \left\{
\begin{aligned}
& \frac{1}{|\mathcal{D}| \cdot n}\sum_{i\in \mathcal{D}} \sum_{k=0}^{n-1} \left( 
M_i^k - {M_i^{k}}^\prime \right)^2 & if\quad BitACC < \tau_3 \\
& \frac{1}{|\mathcal{D}|}\sum_{i\in \mathcal{D}} \log \left( \sum_{k=0}^{n-1} \exp{(M_i^k - {M_i^{k}}^\prime)^2} \right) & otherwise
\end{aligned}
\right.$
}
\end{equation}
where $BitACC$ represents Bit Accuracy and $\tau_3$ is a threshold. When Bit Accuracy is high, the regression loss for bits is small and Eq.\ref{eq:step4-bit loss} can help enhance the magnitude of the supervision signal. Finally, the loss function of Step 3 can be expressed by Eq.\ref{eq:loss L4}:
\begin{equation}
\label{eq:loss L4}
    \mathcal{L}_{3} = \alpha_1 \mathcal{L}_z + \alpha_2 \mathcal{L}_I + \alpha_3 \mathcal{L}_M
\end{equation}
where $\alpha_1$, $\alpha_2$ and $\alpha_3$ are the weighting coefficients. The latent diffusion model is frozen throughout the three steps.

\begin{table*}[htbp]
    \footnotesize
    \centering
    \renewcommand\arraystretch{1.2}
    \caption{The results on image quality and watermark robustness for previous methods and ours. D. Avg.: the average results on the destructive attacks. C. Avg.: the average results on the constructive attacks. R. Avg.: the average results on the reconstructive attacks. Avg.: the average results on all the attacks. B.: Bit Accuracy (\%). T.: TPR@0.01FPR (\%). S.Signa.: StableSignature. S.Stamp: StegaStamp. DT: DiffuseTrace. The numbers in parentheses are the bit lengths. \colorbox{lightgray}{Mark} indicates that the metrics of our method are better than or equal to the ones of the previous methods. \underline{Underline} indicates the best results of the previous methods.}
    \begin{tabular}{p{1.8cm}<{\centering} | p{0.7cm}<{\centering} p{0.7cm}<{\centering}  p{0.7cm}<{\centering} p{0.7cm}<{\centering} p{0.7cm}<{\centering}  p{0.7cm}<{\centering} | p{0.6cm}<{\centering} p{0.6cm}<{\centering} p{0.6cm}<{\centering} p{0.6cm}<{\centering} p{0.6cm}<{\centering} p{0.6cm}<{\centering} | p{0.6cm}<{\centering}  p{0.6cm}<{\centering} }
    \toprule
    \multirow{2}{*}{Method} & \multicolumn{6}{c|}{Untouched watermarked images} & \multicolumn{2}{c}{D. Avg.} & \multicolumn{2}{c}{C. Avg.} & \multicolumn{2}{c|}{R. Avg.} & \multicolumn{2}{c}{Avg.}  \\ 
     & FID$\downarrow$ & SSIM$\uparrow$ & NIQE$\downarrow$ & PIQE$\downarrow$ & B. & T. & B. & T. & B. & T. & B. & T. & B. & T.  \\
    \midrule
    \multicolumn{15}{c}{MS-COCO 2017 Evaluation \cite{lin2014coco}} \\
    \midrule
    \tabincell{c}{TreeRing} & 26.38 & 0.02 & 13.05 & 11.95 & - & 97.40 & - & 48.30 & - & 43.77 & - & 38.20 & - & 44.67  \\
    \tabincell{c}{S.Signa. (48)} & 25.75 & 0.92 & \underline{12.77} & 10.98 & 99.45 &  \underline{100.0} & 58.65 & 25.71 & 68.19 & 49.36 & 59.83 & 34.82& 60.91 & 33.17 \\
    \tabincell{c}{LaWa (48)} & \underline{24.72} & 0.93 & 12.82 & \underline{10.92} & \underline{99.92} & \underline{100.0} & 84.07 & 74.55 & 87.28 & 88.94 & 81.96 & 82.57 & 84.08 & 79.83\\
    \tabincell{c}{RoSteALS (56)} & 35.60 & 0.93 & 15.81 & 44.35 & 96.69 & 94.69 & \underline{84.84} & \underline{88.45} & \underline{96.57} & \underline{95.25} & 85.10 & 74.07 & 87.26 & 85.50  \\
    \tabincell{c}{S.Stamp (56)} & 25.37 & \underline{0.95} & 16.91 & 11.33 & 96.98 & 94.69 & 83.86 & 78.11 & 96.54 & 95.21 & \underline{94.27} & \underline{95.33} & \underline{89.52} & \underline{86.69} \\
    \tabincell{c}{DT (56)} & 31.86 & 0.00 & 15.39 & 18.74 & 58.03 & 16.42 & 54.05 & 8.14 & 56.60 & 13.05 & 56.51 & 12.21 & 55.30 & 10.34  \\
    \midrule
    
    Ours (48) & \cellcolor{lightgray!100}{24.59} & \cellcolor{lightgray!100}{0.97} & \cellcolor{lightgray!100}{12.12} & \cellcolor{lightgray!100}{10.11} & \cellcolor{lightgray!100}{99.93} & \cellcolor{lightgray!100}{100.0} & \cellcolor{lightgray!100}{91.66} & \cellcolor{lightgray!100}{98.29} & \cellcolor{lightgray!100}{98.78} & \cellcolor{lightgray!100}{100.0} & \cellcolor{lightgray!100}{98.10} & \cellcolor{lightgray!100}{99.99} & \cellcolor{lightgray!100}{95.02} & \cellcolor{lightgray!100}{99.14} \\
    Ours (56) & \cellcolor{lightgray!100}{24.53} & \cellcolor{lightgray!100}{0.95} & \cellcolor{lightgray!100}{12.33} & \cellcolor{lightgray!100}{9.86} & 99.89 & \cellcolor{lightgray!100}{100.0} & \cellcolor{lightgray!100}{92.41} & \cellcolor{lightgray!100}{96.70} & \cellcolor{lightgray!100}{98.94} & \cellcolor{lightgray!100}{100.0} & \cellcolor{lightgray!100}{98.20} & \cellcolor{lightgray!100}{100.0} & \cellcolor{lightgray!100}{95.45} & \cellcolor{lightgray!100}{98.35} \\
    Ours (64) & \cellcolor{lightgray!100}{24.59} & \cellcolor{lightgray!100}{0.95} & \cellcolor{lightgray!100}{12.33} & \cellcolor{lightgray!100}{10.01} & \cellcolor{lightgray!100}{99.95} & \cellcolor{lightgray!100}{100.0} & \cellcolor{lightgray!100}{92.13} & \cellcolor{lightgray!100}{97.45} & \cellcolor{lightgray!100}{98.49} & \cellcolor{lightgray!100}{100.0} & \cellcolor{lightgray!100}{98.19} & \cellcolor{lightgray!100}{99.99} & \cellcolor{lightgray!100}{95.22} & \cellcolor{lightgray!100}{98.72} \\
    Ours (128) & \cellcolor{lightgray!100}{24.59} & \cellcolor{lightgray!100}{0.96} & \cellcolor{lightgray!100}{12.16} & \cellcolor{lightgray!100}{10.14} & 99.45 & \cellcolor{lightgray!100}{100.0} & \cellcolor{lightgray!100}{85.94} & \cellcolor{lightgray!100}{96.38} & 96.24 & \cellcolor{lightgray!100}{100.0} & 93.66 & \cellcolor{lightgray!100}{99.96} & \cellcolor{lightgray!100}{90.32} & \cellcolor{lightgray!100}{98.18} \\

    \midrule
    \midrule
    \multicolumn{15}{c}{Flickr30k Evaluation \cite{young2014image}} \\
    \midrule
    \tabincell{c}{TreeRing} & 37.03 & 0.02 & 13.48 & 10.42 & - & 89.10 & - & 49.26 & - & 40.59 & - & 42.74 & - & 45.57 \\
    \tabincell{c}{S.Signa. (48)} & 36.89 & 0.93 & 13.15 & 10.52 & 99.69 & \underline{100.0} & 58.62 & 25.51 & 68.84 & 50.10 & 59.99 & 36.51 & 61.07 & 33.73 \\
    \tabincell{c}{LaWa (48)} & \underline{35.08} & 0.93 & \underline{12.76} & \underline{9.58} & \underline{99.92} & \underline{100.0} &  81.30 & 68.56 & 82.79 & 80.91 & 76.75 & 75.49 & 80.23 & 73.11\\
    \tabincell{c}{RoSteALS (56)} & 42.01 & 0.93 & 16.31 & 44.98 & 96.49 & 94.60 & \underline{84.80} & \underline{88.37} & 92.90 & 88.22 & 93.37 & \underline{94.68} & 88.99 & \underline{90.23} \\
    \tabincell{c}{S.Stamp (56)} & 36.70 & \underline{0.95} & 17.49 & 11.01 & 96.96 & 94.66 & 84.45 & 79.68 & \underline{96.56} & \underline{95.09} & \underline{94.21} & 94.63 & \underline{89.80} & 87.25  \\
    \tabincell{c}{DT (56)} & 44.14 & 0.01 & 15.47 & 18.60 & 57.82 & 16.11 & 53.83 & 8.10 & 55.34 & 11.17 & 56.13 & 12.09 & 54.82 & 9.91 \\
    \midrule
    Ours (48) & 35.52 & \cellcolor{lightgray!100}{0.97} & \cellcolor{lightgray!100}{12.43} & \cellcolor{lightgray!100}{9.57} & \cellcolor{lightgray!100}{99.96} & \cellcolor{lightgray!100}{100.0} & \cellcolor{lightgray!100}{93.62} & \cellcolor{lightgray!100}{98.34} & \cellcolor{lightgray!100}{98.44} & \cellcolor{lightgray!100}{100.0} & \cellcolor{lightgray!100}{98.73} & \cellcolor{lightgray!100}{99.98} & \cellcolor{lightgray!100}{96.12} & \cellcolor{lightgray!100}{99.16} \\
    Ours (56) & 35.92 & \cellcolor{lightgray!100}{0.96} & \cellcolor{lightgray!100}{12.66} & \cellcolor{lightgray!100}{9.19} & \cellcolor{lightgray!100}{99.92} & \cellcolor{lightgray!100}{100.0} & \cellcolor{lightgray!100}{93.78} & \cellcolor{lightgray!100}{97.00} & \cellcolor{lightgray!100}{98.48} & \cellcolor{lightgray!100}{100.0} & \cellcolor{lightgray!100}{98.85} & \cellcolor{lightgray!100}{100.0} & \cellcolor{lightgray!100}{96.24} & \cellcolor{lightgray!100}{98.50} \\
    Ours (64) & 35.49 & \cellcolor{lightgray!100}{0.95} & \cellcolor{lightgray!100}{12.64} & \cellcolor{lightgray!100}{9.40} & \cellcolor{lightgray!100}{99.97} & \cellcolor{lightgray!100}{100.0} & \cellcolor{lightgray!100}{92.70} & \cellcolor{lightgray!100}{97.65} & \cellcolor{lightgray!100}{98.39} & \cellcolor{lightgray!100}{100.0} & \cellcolor{lightgray!100}{98.74} & \cellcolor{lightgray!100}{99.99} & \cellcolor{lightgray!100}{95.65} & \cellcolor{lightgray!100}{98.82}  \\
    Ours (128) & 35.30 & \cellcolor{lightgray!100}{0.97} & \cellcolor{lightgray!100}{12.44} & \cellcolor{lightgray!100}{9.52} & 99.84 & \cellcolor{lightgray!100}{100.0} & \cellcolor{lightgray!100}{86.54} & \cellcolor{lightgray!100}{96.40} & 96.12 & \cellcolor{lightgray!100}{100.0} & \cellcolor{lightgray!100}{94.94} & \cellcolor{lightgray!100}{100.0} & \cellcolor{lightgray!100}{90.98} & \cellcolor{lightgray!100}{98.20}  \\

    \bottomrule
    \end{tabular}

    \label{tab:attack}
\end{table*}

\section{Experiments, Results and Discussions}
\label{sec: experiments}

\subsection{Experiment Setup}
\label{sec:setup}
\subsubsection{Datasets} 
Three datasets are used to train and evaluate LW. For training LW, 50,000 images are used. They are all randomly sampled from LAION-Aesthetics-5+ \cite{schuhmann2022laion}, which has 600M image-text pairs with predicted aesthetics scores of 5 or higher and is the training set for Stable Diffusion \cite{rombach2022high}. For evaluating LW and other methods, we randomly sample 5,000 captions from two datasets respectively, i.e. MS-COCO 2017 (COCO) \cite{lin2014coco} and Flickr30k \cite{young2014image}, to generate images. The captions from COCO are all sampled in the evaluation subset.

\subsubsection{Watermark Baselines}
Six recently proposed watermarking methods are used as our baselines for comparison, i.e. StegaStamp \cite{tancik2020stegastamp}, StableSignature \cite{fernandez2023stable}, RoSteALS \cite{bui2023rosteals}, LaWa \cite{rezaei2024lawa}, TreeRing \cite{wen2023tree}, and DiffuseTrace \cite{lei2024diffusetrace}. For StegaStamp, StableSignature, RoSteALS, and LaWa, the official codes and checkpoints are used to report the results. The lengths of message bits for the four methods are 56, 48, 56, and 48 respectively. For TreeRing, we follow the settings suggested by \cite{wen2023tree} using the official codes, and inject the watermark with a ring radius of 10. For DiffuseTrace, we set the length of message bits as 56 bits and use the official codes and configurations to train the models.

\subsubsection{LW Structure}
For the message coupler, a U-net is used to fuse latent images and encoded messages. It has the same structure as the noise predictor $\epsilon_\theta$ in Stable Diffusion except for the cross-attention modules. For the message decoupler, another U-net which has the same structure with the coupler is used to decouple encoded messages from latent images. The default methods are used to initialize the parameters of the decoupler. For the message encoder, four fully connected layers are used to map messages into the latent space. $Tanh(\cdot)$ is used as the activation function. For the message decoder, four fully connected layers are used to map decoupled latent images into the form of bits. And $Tanh(\cdot)$ is also used as the activation function. For outputs of the decoder, the values greater than 0 are assigned 1 and the values less than 0 are assigned 0.

\subsubsection{LW Training}
AdamW \cite{loshchilov2018decoupled} is used to optimize the model. For the message encoder and decoder, the learning rate in Step 1 is set to 1e-4 and in Step 3 is 1e-5. For the message coupler and decoupler, the learning rate in Step 2 is set to 1e-3 and in Step 3 is 1e-5. $\tau_1$, $\tau_2$ and $\tau_3$ are set to 0.990, 0.045 and 0.900. $\alpha_1$, $\alpha_2$ and $\alpha_3$ are set to 1.5, 1.0 and 1.0 respectively. The batch size is 2. In Step 3, after the exponential moving average of Bit accuracy reaches 99.00\%, we proceed to train the model for one more epoch and then terminate the training process. No data augmentation or noise layer is used during training. The latent diffusion model used in the experiments is Stable Diffusion v1.4. DDIM sampler \cite{song2020denoising} is used in the generation process and the sampling step is 50.

\subsubsection{Attack Methods}
The following 10 attacks are used for robustness evaluation in our experiments.
\begin{itemize}
    \item Destructive attacks: Brightness distortion, contrast distortion, JPEG compression, Gaussian noising, and Cropping.
    \item Constructive attacks: Gaussian denoising and BM3D denoising \cite{dabov2007image}.
    \item Reconstructive attacks \cite{zhao2023generative}: SD (v2.1) \cite{rombach2022high}, VAE-Cheng \cite{cheng2020learned} and VAE-BMSHJ \cite{balle2018variational}.
\end{itemize}

The distortion factors of brightness and contrast are set to 0.1. The quality factor of JPEG is set to 0.1. The standard deviation of Gaussian Noising is set to 1.0. These options are significantly more challenging than existing studies \cite{tancik2020stegastamp, bui2023rosteals, wen2023tree, fernandez2023stable, zhao2023generative}. For Cropping\&Resizing, each image is cropped proportionally along its side and re-scaled to its original size. The proportion is set to 50\%. When the proportion is above 50\%, the performance of all the methods drops drastically and we cannot distinguish them effectively. Actually, excessive crops can seriously destroy images hence they do not often happen in daily use. For Gaussian Denoising, the kernel width is set to 9. For BM3D, the standard deviation of noise is set to 0.9. When the parameters of Gaussian Denoising and BM3D increase, the performance of the methods remains stable. For the reconstructive attacks, we use the parameters recommended by \cite{zhao2023generative}.

\subsubsection{Evaluation Metrics} 
For image quality, two reference-based metrics, i.e. Frechet Inception Distance (FID) and Structural Similarity Index Measure (SSIM), and two no-reference metrics, i.e. Natural Image Quality Evaluator score (NIQE) \cite{Anish2013niqe} and Perception-based Image Quality Evaluator score (PIQE) \cite{venkatanath2015piqe}, are used. FID measures the distribution difference between 5,000 real images and 5,000 generated images. The real images are from COCO or Flickr30k and have the corresponding captions used for generation.  SSIM measures the similarity between watermarked images and vanilla ones. NIQE and PIQE measure the perceptual quality of images based on the statistical properties of natural images and human vision systems respectively. Vanilla images refer to the generated images by the same prompts with the same diffusion model but without using watermarking methods. Image quality is evaluated on untouched watermarked images, which suffer no attack. 

For watermark robustness, True Positive Rate at 0.01 False Positive Rate (TPR@0.01FPR) is used to quantify the performance of the identification task, and Bit Accuracy is used to quantify the performance of the attribution task. Following \cite{fernandez2023stable} and \cite{zhao2023generative}, we define TPR@0.01FPR as the identification accuracy using the thresholds when the theoretical FPR is less than 0.01, assuming that the matching or mismatching of each bit between injected and extracted messages can be treated as an i.i.d. variable and obeys the Bernoulli distribution with parameter 0.5. The thresholds for 48, 56, 64, and 128 bits are 33, 38, 42, and 78 bits respectively. Please refer to the supplementary material for more details about it. Bit Accuracy refers to the proportion of correctly extracted bits. We randomly generate binary messages for each bit length, and then they are fixed across the experiments as the ground-truth labels. For StableSignature, the official message is used, which has been fixed by the fine-tuned model. 

We further evaluate the overall performance of image quality and watermark robustness. First, we unify the ranges and perform a positive transformation for FID, NIQE, and PIQE. We calculate the ratio of the metrics of vanilla images to the metrics of watermarked images. Then, we report the average of FID (unified and transformed), SSIM, NIQE (unified and transformed), and PIQE (unified and transformed), and the average of BitACC and TPR@0.01FPR under the three types of attacks. For TreeRing, only TPR@0.01FPR is averaged. The results have been reported in Fig.\ref{fig:compare with other methods}.

\subsection{Main Results}

\begin{table*}[th]
    \footnotesize
    \centering
    \renewcommand\arraystretch{1.2}
    \caption{The detailed results on the attacks on MS-COCO 2017 captions. \colorbox{lightgray}{Mark} indicates that the metrics of our method are better than or equal to the ones of the previous methods. \underline{Underline} indicates the best results of the previous methods.}
    \begin{tabular}{p{1.8cm}<{\centering} |  p{0.7cm}<{\centering}  p{0.8cm}<{\centering} p{0.8cm}<{\centering} p{0.8cm}<{\centering} p{0.7cm}<{\centering} p{0.6cm}<{\centering} | p{0.8cm}<{\centering} p{0.8cm}<{\centering} p{0.6cm}<{\centering} |  p{0.8cm}<{\centering}  p{0.8cm}<{\centering} p{0.8cm}<{\centering} p{0.6cm}<{\centering}}
    \toprule
    \multirow{2}{*}{Method} & \multicolumn{6}{c|}{D. Attack} & \multicolumn{3}{c|}{C. Attack} & \multicolumn{4}{c}{R. Attack} \\
     & Bright & Contrast & JPEG & Noising & Crop & Avg.  & Gaussian & BM3D & Avg. & SD (v2.1) & VAE (Cheng) & VAE (BMSHJ) &  Avg. \\
    \midrule
    \multicolumn{14}{c}{Bit Accuracy} \\
    \midrule
    \tabincell{c}{TreeRing } & - & - & - & - & - & - & - & - & - & - & - & - & - \\
    \tabincell{c}{S.Signa. (48) } & 47.59 & 41.95 & 60.60 &  47.11 & \underline{95.99} & 58.65 & 88.95 & 47.43  & 68.19 & 46.73 & 70.96 & 61.81 & 59.83 \\
    \tabincell{c}{LaWa (48) } & \underline{99.79} & 95.10 & 79.15 & 53.83 & 92.46 & 84.07 & \underline{99.40} & 75.15 & 87.28 & 79.46 & 83.21 & 83.21 & 81.96 \\
    \tabincell{c}{RoSteALS (56) } & 95.29 & \underline{95.58} &  91.67 &  \underline{73.23} & 68.41 & \underline{84.84} &  96.69 & \underline{96.45}  & \underline{96.57} & 68.80 & 92.20 & 94.30 & 85.10 \\
    \tabincell{c}{S.Stamp (56) } & 95.18 & 83.09 & \underline{91.97}  &  63.77 & 85.28 & 83.86 & 97.16  &  95.91  &  96.54& \underline{89.20} & \underline{96.80} & \underline{96.80} & \underline{94.27} \\
    \tabincell{c}{DT (56) } & 57.46 & 57.29 & 55.19 & 50.65 & 49.67 & 54.05 & 58.09 & 55.10 & 56.60 & 57.12 & 56.17 & 56.25&56.51 \\
    \midrule
    Ours (48) & 98.21 & \cellcolor{lightgray!100}{96.96}  & \cellcolor{lightgray!100}{92.81}  & \cellcolor{lightgray!100}{79.16}  & 91.18 & \cellcolor{lightgray!100}{91.66} &  \cellcolor{lightgray!100}{99.90} & \cellcolor{lightgray!100}{97.66}  &  \cellcolor{lightgray!100}{98.78}  & \cellcolor{lightgray!100}{99.09} & \cellcolor{lightgray!100}{97.27} & \cellcolor{lightgray!100}{97.93} & \cellcolor{lightgray!100}{98.10} \\
    Ours (56) &  98.97 & \cellcolor{lightgray!100}{97.31} & \cellcolor{lightgray!100}{92.70} &  \cellcolor{lightgray!100}{79.46} & 93.63 &  \cellcolor{lightgray!100}{92.41} & \cellcolor{lightgray!100}{99.88}  &  \cellcolor{lightgray!100}{97.99} &  \cellcolor{lightgray!100}{98.94}  & \cellcolor{lightgray!100}{99.26} & \cellcolor{lightgray!100}{97.33} & \cellcolor{lightgray!100}{98.01} & \cellcolor{lightgray!100}{98.20} \\ 
    Ours (64) & 98.69 & \cellcolor{lightgray!100}{97.60} &  \cellcolor{lightgray!100}{92.63} &  \cellcolor{lightgray!100}{78.30} & 93.43 & \cellcolor{lightgray!100}{92.13} & \cellcolor{lightgray!100}{99.93}  &  \cellcolor{lightgray!100}{97.05} &  \cellcolor{lightgray!100}{98.49}  & \cellcolor{lightgray!100}{99.19} & \cellcolor{lightgray!100}{97.44} & \cellcolor{lightgray!100}{97.95} & \cellcolor{lightgray!100}{98.19} \\
    Ours (128) & 93.06 & 90.88 & 84.10  & 72.40  & 89.26  &  \cellcolor{lightgray!100}{85.94}  & 97.37   &  95.10 &  96.24  & \cellcolor{lightgray!100}{95.13} & 92.77 & 93.08 & 93.66 \\ 

    \midrule
    \midrule
    \multicolumn{14}{c}{TPR@0.01FPR} \\
    \midrule
    \tabincell{c}{TreeRing} & 73.51 & 65.13 & 23.59  &  3.65 & 75.63 & 48.30 &  82.56 &  4.97 & 43.77 & 52.68 & 31.46 & 33.47 & 38.20 \\
    \tabincell{c}{S.Signa. (48) } & 1.52  & 0.04 &  28.03 & 0.00  & \underline{98.96} & 25.71 & 98.70  &  0.01 & 49.36 & 0.38 & 69.73 & 34.35 & 34.82 \\
    \tabincell{c}{LaWa (48) } & \underline{100.0} & 92.81 & 85.23 & 2.79 & 91.90 & 74.55 & \underline{100.0} & 77.87& 88.94 & 86.39 & 80.64 & 80.69 & 82.57 \\
    \tabincell{c}{RoSteALS (56) } & 95.08 & \underline{95.18} & \underline{94.67}  & \underline{85.23}  & 72.11 & \underline{88.45}  & 94.79  & \underline{95.70}  &  \underline{95.25} & 41.20 & 88.70 & 92.30 & 74.07 \\
    \tabincell{c}{S.Stamp  (56) } & 94.93 & 88.65 &  91.94 & 20.16 & 94.86 & 78.11 & 95.56  & 94.86  & 95.21  & \underline{95.64} & \underline{95.61} & \underline{94.73} & \underline{95.33} \\
    \tabincell{c}{DT (56) } & 13.82 & 14.11 & 10.53 & 1.74 & 0.52&8.14 & 16.79 & 9.3&13.05 & 13.61 & 11.27 & 11.75 & 12.21 \\
    \midrule
    Ours (48) & 99.98 & \cellcolor{lightgray!100}{99.98} & \cellcolor{lightgray!100}{99.94}  & \cellcolor{lightgray!100}{93.28}  & 98.26 & \cellcolor{lightgray!100}{98.29} & \cellcolor{lightgray!100}{100.0}  & \cellcolor{lightgray!100}{100.0}  &  \cellcolor{lightgray!100}{100.0}  & \cellcolor{lightgray!100}{100.0} & \cellcolor{lightgray!100}{99.96} & \cellcolor{lightgray!100}{100.0} & \cellcolor{lightgray!100}{99.99} \\
    Ours (56) & 99.98 & \cellcolor{lightgray!100}{99.46} & \cellcolor{lightgray!100}{99.42}  & \cellcolor{lightgray!100}{85.54}  & \cellcolor{lightgray!100}{99.11} &  \cellcolor{lightgray!100}{96.70}  &   \cellcolor{lightgray!100}{100.0} &  \cellcolor{lightgray!100}{100.0} &  \cellcolor{lightgray!100}{100.0} & \cellcolor{lightgray!100}{100.0} & \cellcolor{lightgray!100}{100.0} & \cellcolor{lightgray!100}{100.0} & \cellcolor{lightgray!100}{100.0} \\
    Ours (64) & \cellcolor{lightgray!100}{100.0} & \cellcolor{lightgray!100}{100.0} & \cellcolor{lightgray!100}{99.98}  & \cellcolor{lightgray!100}{87.36}  & \cellcolor{lightgray!100}{99.92} & \cellcolor{lightgray!100}{97.45} & \cellcolor{lightgray!100}{100.0}  &  \cellcolor{lightgray!100}{100.0} &  \cellcolor{lightgray!100}{100.0}  & \cellcolor{lightgray!100}{100.0} & \cellcolor{lightgray!100}{99.96} & \cellcolor{lightgray!100}{100.0} & \cellcolor{lightgray!100}{99.99} \\
    Ours (128) & \cellcolor{lightgray!100}{100.0}  & \cellcolor{lightgray!100}{100.0} &  \cellcolor{lightgray!100}{99.98} & 84.08  & 97.84 & \cellcolor{lightgray!100}{96.38} & \cellcolor{lightgray!100}{100.0}  & \cellcolor{lightgray!100}{100.0}  & \cellcolor{lightgray!100}{100.0} & \cellcolor{lightgray!100}{100.0} & \cellcolor{lightgray!100}{99.90} & \cellcolor{lightgray!100}{99.98} & \cellcolor{lightgray!100}{99.96} \\

    \bottomrule
    \end{tabular}

    \label{tab: attack coco}
\end{table*}

\begin{figure*}[htb]
  \centering
  \includegraphics[scale=0.29]{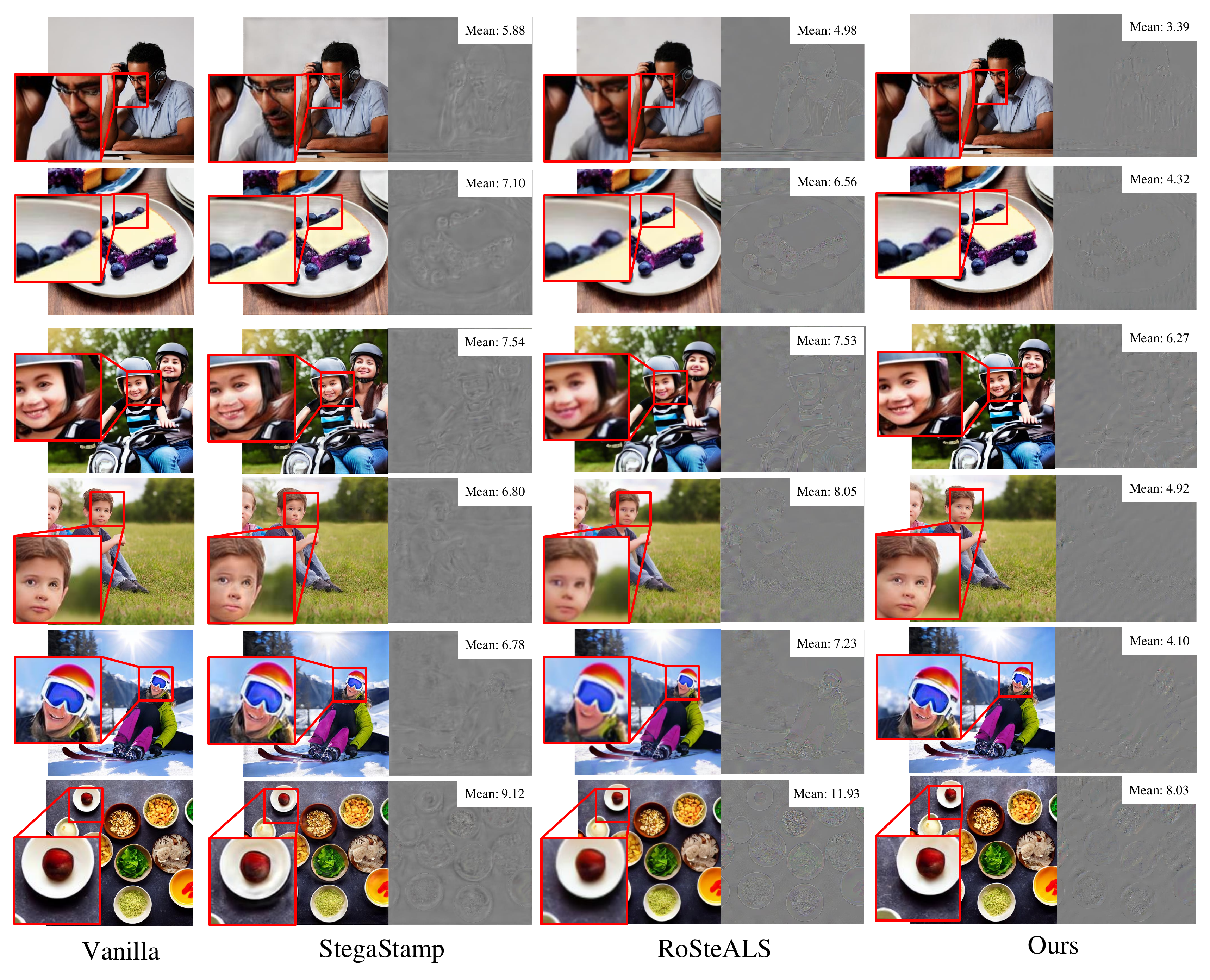}
  \caption{The clean images and the images watermarked by StegaStamp \cite{tancik2020stegastamp} (56 bits), RoSteALS \cite{bui2023rosteals} (56 bits), and LW (56 bits). The residual images between the watermarked images and the clean images are given, with the mean absolute pixel difference in the upper right corner.}
  \label{fig:images}
\end{figure*}

 In Tab.\ref{tab:attack}, we report an overview of image quality and watermark robustness results. From the perspective of image quality, among the compared methods, LaWa excels over other methods in terms of FID and PIQE for both COCO and Flickr captions. It demonstrates that LaWa has an advantage in the perceptual quality of the watermarked images. StableSignature also presents an advantage in the perceptual quality of the watermarked images. StegaStamp achieves superior performance on SSIM but inferior performance on NIQE, indicating its poor perceptual quality. The poor PIQE of RoSteALS on two evaluation sets suggests a large gap between the perceptual quality of the watermarked images and the human vision systems. Besides, its poor FID further demonstrates the significant negative impact on image quality. For TreeRing and DiffuseTrace, their SSIMs are significantly inferior to others because editing the sampled noise changes the content of the generated images. Under the same number of message bits, LW exceeds or is equal to these existing methods on the four metrics in most cases, especially on FID, NIQE, and PIQE. It shows that compared with the previous methods, LW can watermark generated images with the smallest difference and the closest perceptual quality to vanilla ones. When the number of message bits is 64 or even 128, LW is still superior in image quality.

From the perspective of watermark robustness, it can be seen that StegaStamp and RoSteALS have stronger resistance to the attacks than other methods. The robustness of StegaStamp is stronger than RoSteALS for reconstructive attacks, and RoSteALS has an advantage in resisting destructive attacks. When we focus on our proposed LW, under the same (48, 56) or even more number (64) of message bits, the average results on the 10 attacks are all higher than StegaStamp and RoSteALS, at least 5\% higher on Bit Accuracy and 11\% higher on TPR@0.01FPR. When the number of bits is 128, a few BitACC is inferior to these two methods, but the margin is negligible, and its average robustness on all the attacks remains superior to them. The detailed results on each attack for COCO captions are shown in Tab.\ref{tab: attack coco}, and the results for Flickr captions can be found in the supplementary material. Tab.\ref{tab: attack coco} demonstrates that our method has obvious advantages over the compared methods in resisting various attacks, except for Cropping. For Cropping, StableSignature achieves the best performance, while LW lags behind it by nearly 5\% for the same number of bits. However, StableSignature exhibits significantly poor performance under other attacks. As the method closest to ours in image quality, LaWa exhibits poor watermark robustness, especially on the reconstructive attacks.

\begin{figure}[t]
  \centering
  \includegraphics[scale=0.25]{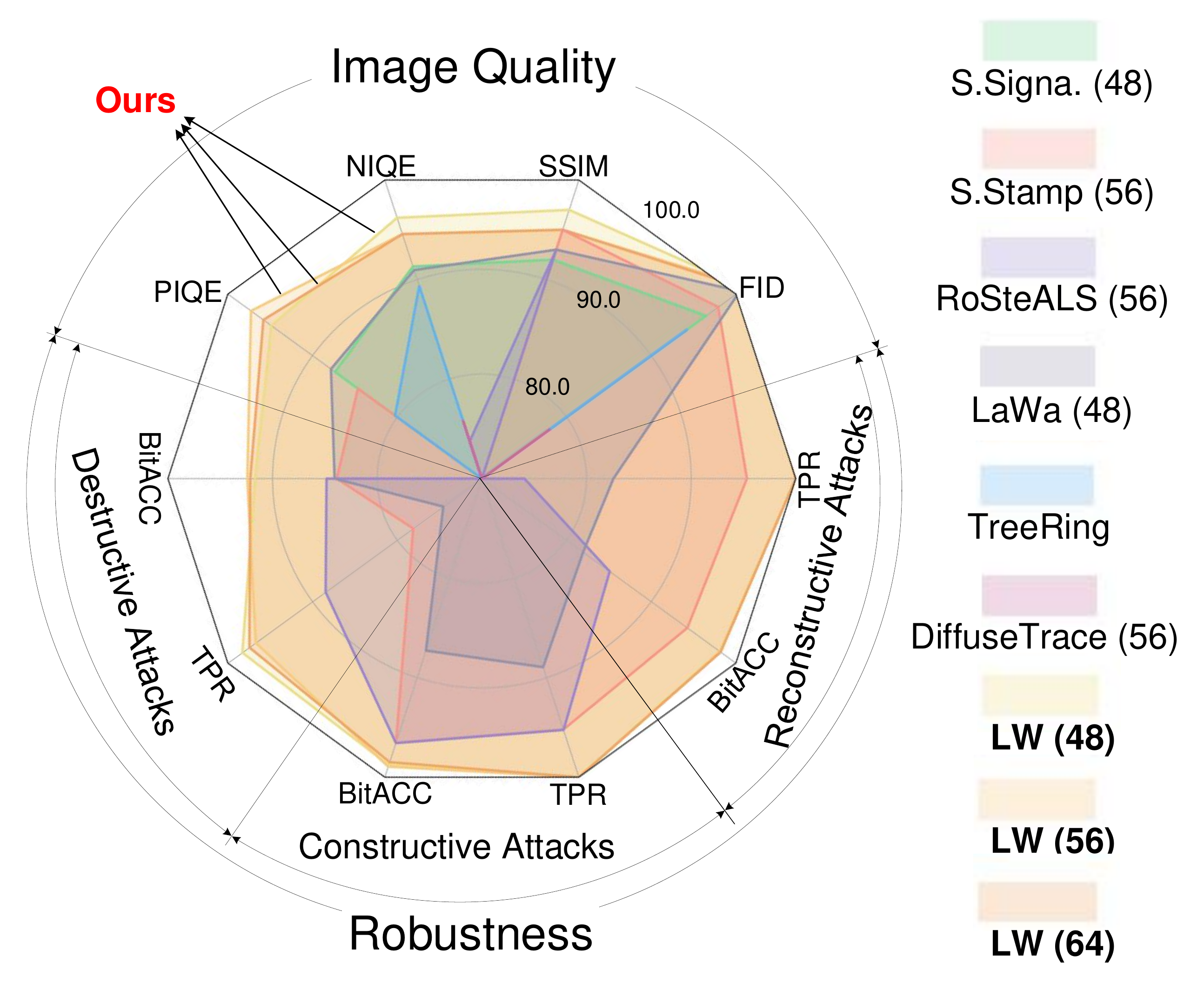}
  \caption{The radar chart for a comprehensive evaluation. The metrics are unified and transformed using the method mentioned in Sec.\ref{sec:setup}.}
  \label{fig: radar}
\end{figure}

Finally, we give some examples of images watermarked by StegaStamp, RoSteALS, and ours in Fig.\ref{fig:images}. StegaStamp and RoSteALS exhibit better robustness compared with other methods. However, it can be seen that the images watermarked by StegaStamp have obvious shadows at the edges and bright areas, and the images watermarked by RoSteALS are blurry while our method does not have these problems. Fig.\ref{fig:images} also shows the residuals and the mean absolute pixel difference between the vanilla and watermarked images, demonstrating LW's image quality closer to the vanilla images. Please refer to the supplementary material for the images watermarked by our method with 48, 56, 64, and 128-bit messages. The examples, along with the radar chart shown in Fig.\ref{fig: radar}, further illustrate the trade-off between image quality and watermark robustness in existing methods, and our proposed LW significantly reduces the contradiction between the two.

\begin{figure*}[t]
  \centering
  \includegraphics[scale=0.12]{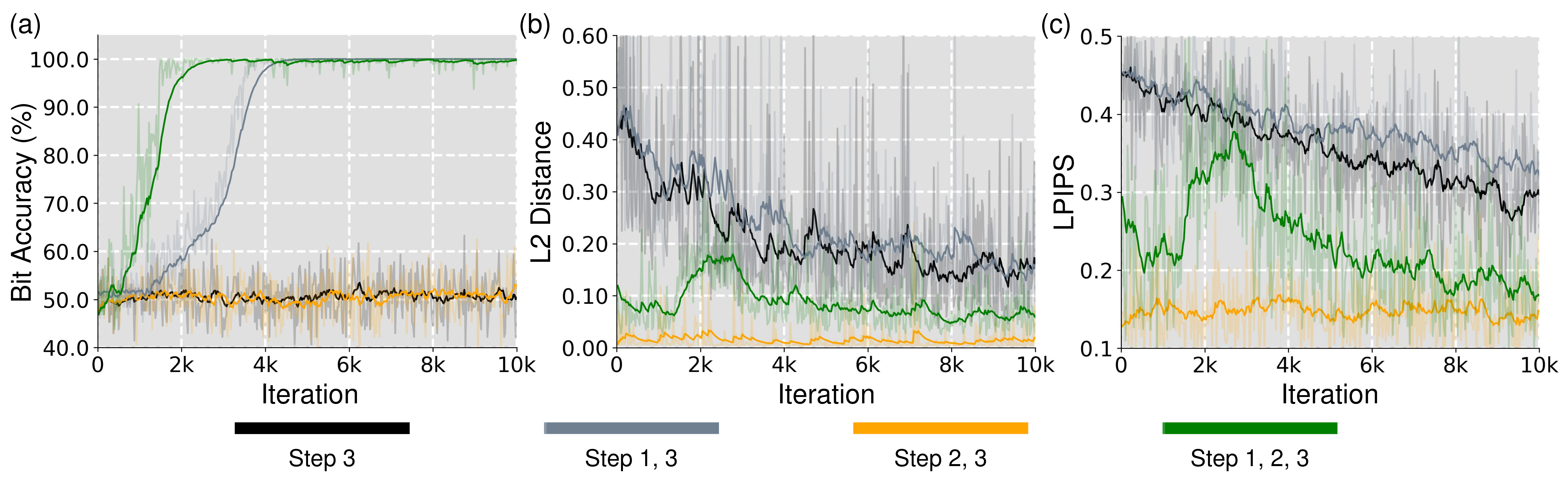}
  \caption{The training curves using one or more of the three steps of the progressive training strategy. (a) Bit Accuracy. (b) L2 distance of latent images (Eq.\ref{eq:step4-latent}). (3) Learned Perceptual Image Patch Similarity (LPIPS) loss \cite{zhang2018unreasonable} (Eq.\ref{eq:step4-lpips}). Step 1 pre-trains the message encoder and decoder. Step 2 pre-trains the message coupler. Step 3 is the formal training.}
  \label{fig:loss}
\end{figure*}

\begin{table*}[t]
    \small
    \centering
    \renewcommand\arraystretch{1.2}
    \captionof{table}{The results of using different latent channel(s) for watermark injection on MS-COCO 2017 captions. The differences between using Channel 0 and other channel(s) are given. \colorbox{red!30}{Mark} denotes the decreasing metrics.}
    \begin{tabular}{p{2.0cm}<{\centering} | p{1.2cm}<{\centering} p{1.2cm}<{\centering}  p{1.2cm}<{\centering} p{1.2cm}<{\centering} p{1.2cm}<{\centering} | p{1.2cm}<{\centering} p{1.2cm}<{\centering}  p{1.2cm}<{\centering} | p{1.2cm}<{\centering} }
    \toprule
    \multirow{2}{*}{Channel(s)} & \multicolumn{5}{c|}{Untouched watermarked images} & D. Avg. & C. Avg. & R. Avg. & Avg. \\ 
     & FID $\downarrow$ & SSIM $\uparrow$ & NIQE $\downarrow$ & PIQE $\downarrow$ & B. & B. & B. & B. & B. \\
    \midrule
    0 (Used) & 24.59 & 0.95 & 12.33 & 10.01 & 99.95 & 92.13 & 98.49 &  98.19 & 96.27 \\
    1 & +0.13 & +0.01 & \cellcolor{red!30}{-0.21} & \cellcolor{red!30}{-0.32} & \cellcolor{red!30}{-0.46} & \cellcolor{red!30}{-0.63} & \cellcolor{red!30}{-0.97} & \cellcolor{red!30}{-1.55} & \cellcolor{red!30}{-0.97} \\
    2 & +0.16 & \cellcolor{red!30}{-0.07} & +0.20 & \cellcolor{red!30}{-0.66} & \cellcolor{red!30}{-0.10} & \cellcolor{red!30}{-0.51} & \cellcolor{red!30}{-1.23} & \cellcolor{red!30}{-2.63} & \cellcolor{red!30}{-1.29} \\
    3 & \cellcolor{red!30}{-0.34} & \cellcolor{red!30}{-0.03} & \cellcolor{red!30}{-0.21} & \cellcolor{red!30}{-0.14} & \cellcolor{red!30}{-0.04} & \cellcolor{red!30}{-0.13} & \cellcolor{red!30}{-0.37} & \cellcolor{red!30}{-1.33} & \cellcolor{red!30}{-0.54}  \\ 
    0,1,2,3 & +0.51 & +0.01 & \cellcolor{red!30}{-0.71} & +0.58 & \cellcolor{red!30}{-0.16} & \cellcolor{red!30}{-0.31} & \cellcolor{red!30}{-0.94} & \cellcolor{red!30}{-2.00} & \cellcolor{red!30}{-0.94}  \\
    \bottomrule
    \end{tabular}
    \label{tab: channels}
\end{table*}  

\subsection{Ablation Studies}
\label{sec:ablation}

\subsubsection{Training Strategy}
\label{sec: strategy}
\noindent To verify the effectiveness of each step in the three-step progressive training strategy, we plot the training curves of Bit Accuracy, $\mathcal{L}_z$ (Eq.\ref{eq:step4-latent}) and $\mathcal{L}_I$ (Eq.\ref{eq:step4-lpips}) using one or more of the training steps in Fig.\ref{fig:loss}. From Fig.\ref{fig:loss} (a), we can see that the message encoder and decoder cannot be optimized without Step 1. Fig.\ref{fig:loss} (b) and (c) show that LW converges faster and the difference between watermarked and vanilla images is smaller with Step 2. The proposed training strategy can help LW obtain watermark injection and detection abilities while retaining image quality better.

\subsubsection{Injection Channel}
\noindent By coupling messages with various channels of latent images, we observe diverse effects on image quality and watermark robustness. In Eq.\ref{eq: channel}, encoded messages are coupled with Channel 0 of latent images. We further evaluate other options and summarize the results in Tab.\ref{tab: channels}. Our experiments indicate that selecting other channels or all channels for watermark injection often hurt its robustness while sometimes it can improve image quality. Considering all the metrics, we recommend selecting Channel 0 for injecting watermarks.

\begin{figure}[t]
    \centering  
    \includegraphics[scale=0.45]{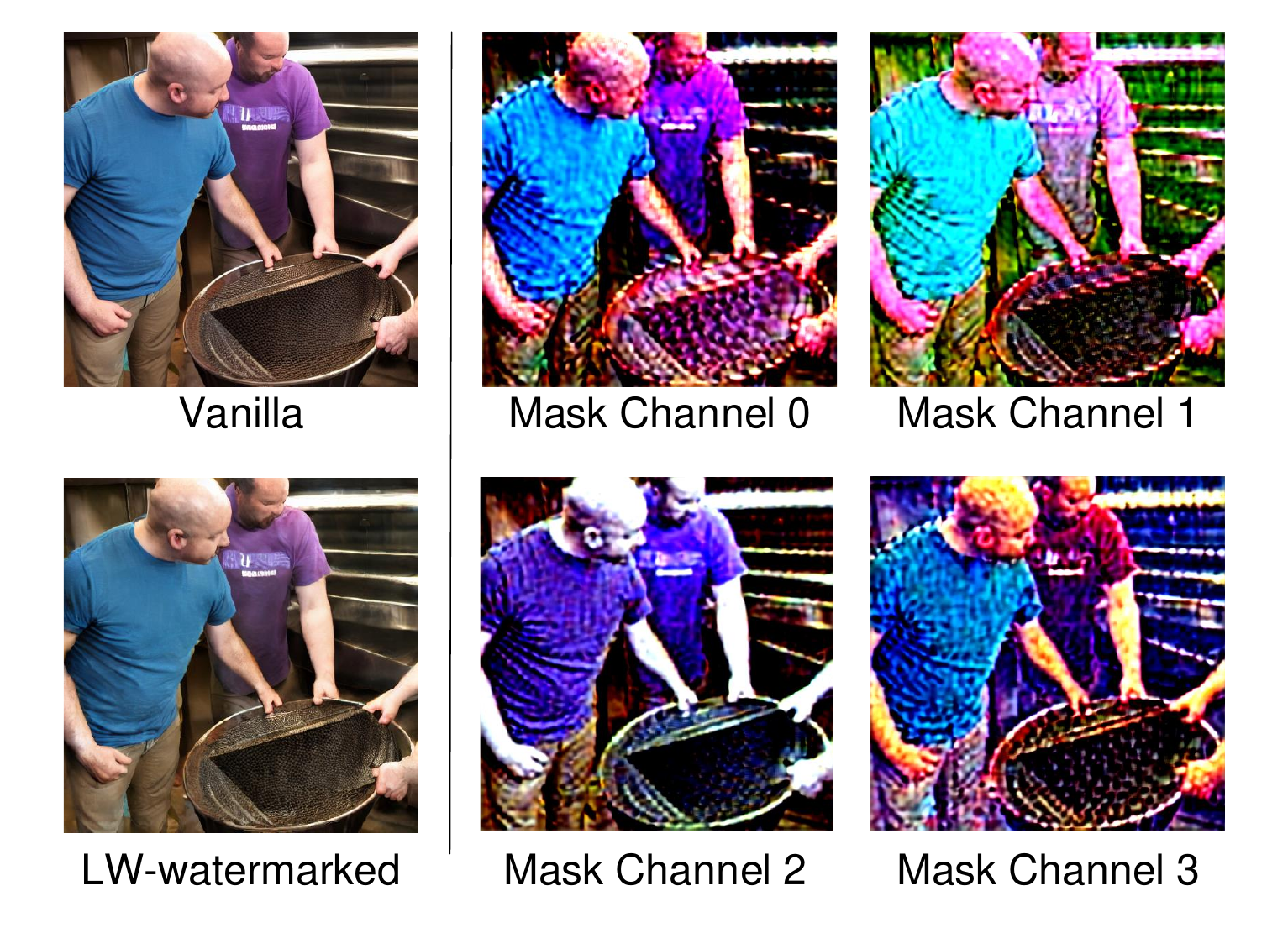} 
    \caption{An example for visualizing the impact of each channel in the latent space on the generated images. On the left, it presents the vanilla-generated image alongside the LW-watermarked one. On the right, it presents the generated images with one latent channel masked. The channel is masked after the diffusion process but before the decoding.}
    \label{fig: latent channel}
\end{figure}

To delve into the reason why selecting different channels brings different performances, we conduct an ablation experiment on the generated images. Specifically, we mask each channel in the latent space respectively after the diffusion process but before the decoding, and then observe the changes in the generated images. Our finding is that masking one channel causes a change in the hues of images, and masking different channels causes different degrees of change, leading to different trade-offs between watermark robustness and image quality. An example is shown in Fig.\ref{fig: latent channel}. In this example, when we mask Channel 1 and Channel 3, the background of the image appears to have distinct green and purple components. When we mask Channel 2, the blue T-shirt turns purple. When we mask Channel 0, overall, its hue is closer to the vanilla image. The above observation reveals that when watermarks with equal strength are injected into the latent space, selecting Channel 0 can ensure maximal consistency with the vanilla image. It explains why injecting watermarks into Channel 0 can achieve better image quality and watermark robustness at the same time. We speculate that the difference in the effects of the channels is formed spontaneously when training the autoencoder, which needs further study.

\subsection{Additional Discussions}

\subsubsection{Necessity of The Progressive Training Strategy }
\label{sec: necessity}
\noindent In this paper, the progressive training strategy is proposed to address the difficulty in training the watermarking-related modules. As mentioned in Sec.\ref{sec: intro}, it stems from the non-adaptation of the watermarking modules and the latent encoder/decoder in the training process. Since the watermark injection and detection are performed entirely in the latent space, we try the latent training strategy to avoid the presence of the latent encoder/decoder. The latent training strategy refers to training the modules entirely in the latent space. The latent images are fed into the message decoupler directly and not decoded into the pixel space during training. All the modules do not undergo any pre-training. The loss $\mathcal{L}_I$ for the pixel images is also removed. The results are shown in Tab.\ref{tab: strategy}. It shows that the latent training strategy significantly degrades both image quality and watermark robustness. The underlying reason lies in the absence of the pixel space during the training process, which creates a gap between the training and inference phases. On the one hand, due to the lack of the visual perception loss, the embedded watermarks exert a more pronounced effect on image quality. On the other hand, the apparent deterioration in image quality renders the watermarks more vulnerable to attacks that target the images themselves. Therefore, it is necessary to introduce the decoder and encoder in the training, which justifies the necessity of the proposed training strategy.

\begin{table}[t]
    \small
    \centering
    \renewcommand\arraystretch{1.2}
    \caption{The results of LW with the latent training strategy and the proposed progressive training strategy (64 bits).}
    \begin{tabular}{p{1.4cm}<{\centering} | p{0.8cm}<{\centering} p{0.8cm}<{\centering}  p{1.2cm}<{\centering} | p{2.0cm}<{\centering} }
    \toprule
    \multirow{2}{*}{Strategy} & \multicolumn{3}{c|}{Untouched} & Attacked \\ 
     & FID$\downarrow$ & SSIM$\uparrow$ & BitACC$\uparrow$ & Avg. BitACC$\uparrow$ \\
    \midrule
    Latent & 43.02 & 0.83 & 99.56 & 88.83 \\
    Progressive & \textbf{24.59} & \textbf{0.95} & \textbf{99.95} & \textbf{95.22} \\
    \bottomrule
    \end{tabular}
    \label{tab: strategy}
\end{table}

\subsubsection{Training Size}
\begin{figure*}[t]
    \centering  
    \includegraphics[scale=0.45]{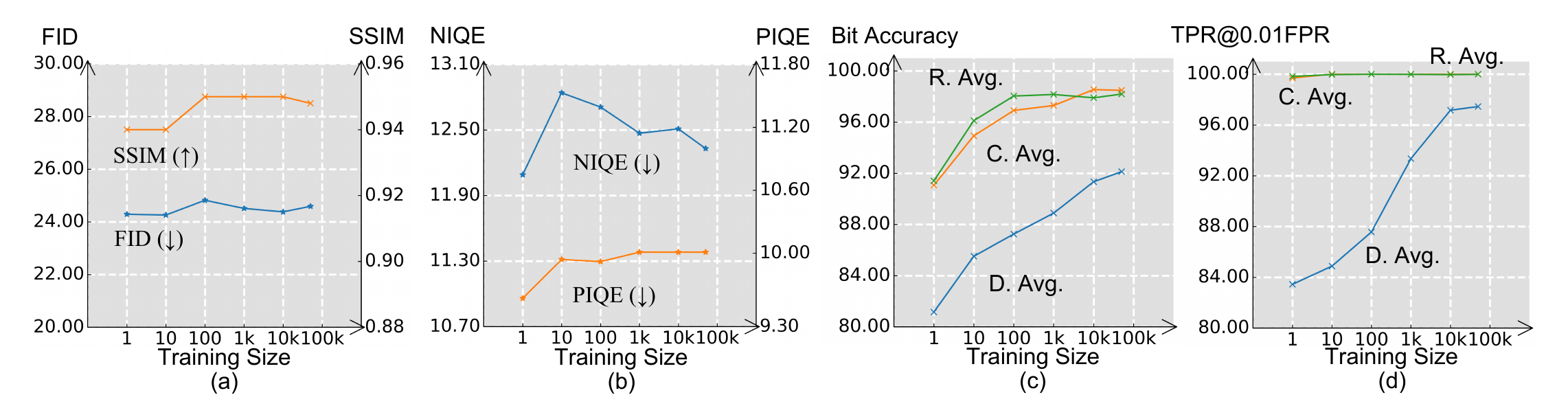} 
    \caption{The change of performance with respect to the size of training data. (a) The reference-based metrics for image quality: FID and SSIM. (b) The no-reference metrics for image quality: NIQE and PIQE. (c) The metrics for watermark robustness: Bit Accuracy under the three attack categories. (d) The metrics for watermark robustness: TPR@0.01FPR under the three attack categories.}  
    \label{fig: training size}
\end{figure*}

\noindent Fig.\ref{fig: training size} illustrates the performance of image quality and watermark robustness when training LW with varying training sizes of 1, 10, 100, 1k, 10k, and 50k images. Notably, the training size has a more significant influence on NIQE, BitACC, and TPR@0.01FPR compared to other metrics. They increase rapidly as the training size increases, while the rest of the metrics change slowly. When the size reaches 50k, the change in most metrics becomes relatively minor, leading us to select this size for training LW. It is worth mentioning that the size of the training data employed in this work is smaller than the 100k images used in RoSteALS \cite{bui2023rosteals}. StegaStamp \cite{tancik2020stegastamp} and StableSignature \cite{fernandez2023stable} do not provide specific details regarding the size of the training data from MIRFLICKR \cite{huiskes2008mirflickr} and MS-COCO \cite{lin2014coco} that they use respectively.

\subsubsection{Results on Vanilla Images}
\label{sec: vanilla images}
\noindent We evaluate the False Positive Rate (FPR) of LW using the thresholds for TPR@0.01FPR. The evaluation is conducted on the vanilla images generated by the Stable Diffusion Model with the same prompts used in the evaluation on MS-COCO 2017 \cite{lin2014coco} and Flickr30k \cite{young2014image}. The results are shown in Tab.\ref{tab: fpr}. We also report the corresponding theoretical values under the thresholds. Tab.\ref{tab: fpr} shows that the FPR of LW is very close to the theoretical values under various lengths of message bits and is lower than $0.01$ as well. It demonstrates that the theoretical assumptions and calculation methods of TPR@0.01FPR are suitable for LW.

\begin{table}[t]
    \centering
    \caption{The results of LW on the vanilla images. False Positive Rate (FPR, \%) are reported with the corresponding theoretical values. The thresholds are the ones used to calculate TPR@0.01FPR in the experiments. The vanilla images are generated by the Stable Diffusion with the same prompts in the evaluation on MS-COCO 2017 \cite{lin2014coco} and Flickr30k \cite{young2014image}.}
    \begin{tabular}{p{1.5cm}<{\centering} | p{1.5cm}<{\centering} | p{1.5cm}<{\centering} p{1.5cm}<{\centering}}
    \toprule
    \# of Bits & Theoretical & COCO & Flickr30k\\ 
    \midrule
    48 & 0.0066 & 0.0074  & 0.0074 \\
    56 & 0.0052 & 0.0056  & 0.0056 \\
    64 & 0.0084 & 0.0080  & 0.0080 \\
    128 & 0.0083 & 0.0072  & 0.0078 \\
    \bottomrule
    \end{tabular}
    \label{tab: fpr}
\end{table}

\subsubsection{Time Efficiency}
In addition to a low training cost and high performance, a good watermarking method should also have a low time cost for watermark injection and extraction. In this section, we measure the time efficiency of the methods. The results are averaged over 5,000 images with a size of $512\times 512$ and reported in Tab.\ref{tab: time}.  The inference time of LW ranks second, which is faster than most of the compared methods.

\begin{table}[t]
    \small
    \centering
    \renewcommand\arraystretch{1.2}
    \caption{The inference time of the methods. The results are averaged over 5,000 images. The numbers in parentheses are the bit lengths. \textbf{Bold} results indicate the best ones and \underline{underline} results indicate the second ones.}
    \begin{tabular}{p{2.5cm}<{\centering} | p{1.6cm}<{\centering}  p{1.6cm}<{\centering} p{1.6cm}<{\centering}}
    \toprule

    \multirow{2}{*}{Method} & \multicolumn{3}{c}{Time (millisecond)} \\
    & Injection & Extraction & Total \\ \midrule
    
    S.Signa. (48) & \textbf{0.0} & \underline{43.2} & \textbf{43.2} \\
    S.Stamp (56) & 140.2 & 168.2 & 308.4 \\
    RoSteALS (56) & 305.0 & 76.9 & 381.9 \\
    LaWa (48) &  118.9 & \textbf{11.6} &  130.5 \\
    TreeRing & 17.7 & 1176.3 & 1194.0\\
    DiffuseTrace (56) &  \underline{2.2} & 3926.8  & 3929.0\\
    LW (56) & 67.8 & 60.1 & \underline{127.9} \\

    \bottomrule
    
    \end{tabular}
    
    \label{tab: time}
\end{table}

\subsubsection{Limitation}
LW lacks rotation invariance, which is a limitation of our method. While it retains robustness to minor image rotations, LW fails to effectively resist large rotations. To illustrate it, we conduct the experiments by rotating images at varying angles and comparing the extraction performance of various methods. Our findings reveal that for a rotation angle of 5\degree, LW (56 bits) achieves a BitACC exceeding 95\%. However, as the angle surpasses 10\degree, the BitACC of LW (56 bits) drops below 60\%, which outperforms RoSteALS and StegaStamp but trails behind StableSignature. Notably, large rotations can be easily detected and rectified, offering a potential pretext defense mechanism to mitigate the shortcomings of existing methods, including ours. We leave the exploration of watermarking methods in the latent diffusion space with rotation invariance as future work.

\section{Conclusion}
\noindent In this paper, we propose LW with a progressive training strategy to root watermarks in latent space for latent diffusion models, to identify and attribute generated images. Compared with previous methods, the fundamental difference is that watermarks are injected and detected in latent diffusion space rather than pixel space. It weakens the direct correlation between image quality and watermark robustness and alleviates their contradiction. We evaluate the image quality and robustness against 10 attacks of LW as well as six previous methods on two datasets, demonstrating that the proposed method brings stronger robustness and higher image quality at the same time. We provide an effective tool for distinguishing and attributing synthetic images in the AIGC era.

\bibliographystyle{unsrtnat}
\bibliography{main}  

\clearpage

\setcounter{section}{0}
\setcounter{table}{0}
\setcounter{figure}{0}
\setcounter{equation}{0}
\renewcommand\thesection{Appendix \Alph{section}}
\renewcommand{\theequation}{S\arabic{equation}}
\renewcommand{\thefigure}{S\arabic{figure}}
\renewcommand{\thetable}{S\arabic{table}}

\section{Detailed Results on Flickr30k}
The results on Flickr30k captions under the single-attack scenarios are shown in Tab.\ref{tab: supp attack flickr}. 

\begin{table*}[h]
    \scriptsize
    \centering
    \renewcommand\arraystretch{1.2}
    \caption{The detailed results on the attacks on Flickr30k captions. D. Attack: the destructive attacks. C. Attack: the constructive attacks. R. Attack: the reconstructive attacks. S.Signa.: StableSignature. S.Stamp: StegaStamp. The numbers in parentheses are the numbers of encoded bits. \colorbox{lightgray}{Mark} indicates that the B. and T. of our method are better than or equal to the ones of the previous methods. \underline{Underline} indicates the best results of the previous methods.}
    \begin{tabular}{p{1.6cm}<{\centering} |  p{0.7cm}<{\centering}  p{0.8cm}<{\centering} p{0.8cm}<{\centering} p{0.8cm}<{\centering} p{0.7cm}<{\centering} p{0.6cm}<{\centering} | p{0.8cm}<{\centering} p{0.8cm}<{\centering} p{0.6cm}<{\centering} |  p{0.8cm}<{\centering}  p{0.8cm}<{\centering} p{0.8cm}<{\centering} p{0.6cm}<{\centering} }
    \toprule
    \multirow{2}{*}{Method} & \multicolumn{6}{c|}{D. Attack} & \multicolumn{3}{c|}{C. Attack} & \multicolumn{4}{c}{R. Attack} \\
     & Bright & Contrast & JPEG & Noising & Crop & Avg.  & Gaussian & BM3D & Avg. & SD (v2.1) & VAE (Cheng) & VAE (BMSHJ) & Avg. \\
    \midrule
    \multicolumn{14}{c}{Bit Accuracy} \\
    \midrule
    \tabincell{c}{TreeRing } & - & - & - & - & - & - & - & - & - & - & - & - & - \\
    \tabincell{c}{S.Signa. (48) } & 46.71 & 41.66 & 60.66 & 47.10  & \underline{96.97} & 58.62 & 89.74 &  47.94 & 68.84 & 45.72 & 71.44 & 62.80 & 59.99 \\
    \tabincell{c}{LaWa (48)} & 92.62 & 91.60 & 76.73 & 53.49 & 92.07 & 81.30 & 92.07 & 73.51 & 82.79 & 71.77 & 81.48 & 77.00 & 76.75 \\
    \tabincell{c}{RoSteALS (56) } & 94.98 & \underline{95.24} & 91.61  & \underline{73.85}  & 68.33 & \underline{84.80} & 96.49  & 89.30  & 92.90 & \underline{90.32} & 94.77 & 95.03 & 93.37 \\
    \tabincell{c}{S.Stamp (56) } & \underline{95.37} & 83.69 & \underline{93.74}  &  64.40 & 85.04 & 84.45 & \underline{97.17}  &  \underline{95.95}  & \underline{96.56} & 88.91 & \underline{96.83} & \underline{96.89} & \underline{94.21} \\
    \tabincell{c}{DT (56)} & 56.65 & 56.45 & 55.08 & 50.71 & 50.26 & 53.83 & 56.42 & 54.25 & 55.34 & 56.57 & 56.02 & 55.80 & 56.13 \\
    \midrule
    Ours (48) & \cellcolor{lightgray!100}{99.71} & \cellcolor{lightgray!100}{97.73}  &  \cellcolor{lightgray!100}{94.09} & \cellcolor{lightgray!100}{80.31}  & 96.26 & \cellcolor{lightgray!100}{93.62} &  \cellcolor{lightgray!100}{99.94} & \cellcolor{lightgray!100}{96.93}  &  \cellcolor{lightgray!100}{98.44}  & \cellcolor{lightgray!100}{99.22} & \cellcolor{lightgray!100}{98.31} & \cellcolor{lightgray!100}{98.67} & \cellcolor{lightgray!100}{98.73} \\
    Ours (56) &  \cellcolor{lightgray!100}{99.01} & \cellcolor{lightgray!100}{98.49} &  \cellcolor{lightgray!100}{94.96} &  \cellcolor{lightgray!100}{79.69} & 96.73 & \cellcolor{lightgray!100}{93.78}  & \cellcolor{lightgray!100}{99.44}  & \cellcolor{lightgray!100}{97.51}  &  \cellcolor{lightgray!100}{98.48}  & \cellcolor{lightgray!100}{99.14} & \cellcolor{lightgray!100}{98.74}  & \cellcolor{lightgray!100}{98.68} & \cellcolor{lightgray!100}{98.85} \\
    Ours (64) & \cellcolor{lightgray!100}{99.06}  & \cellcolor{lightgray!100}{98.16}  &  \cellcolor{lightgray!100}{93.83} &  \cellcolor{lightgray!100}{78.13}  & 94.34 & \cellcolor{lightgray!100}{92.70}  &  \cellcolor{lightgray!100}{99.95}  & \cellcolor{lightgray!100}{96.83}  &  \cellcolor{lightgray!100}{98.39}  & \cellcolor{lightgray!100}{99.32} & \cellcolor{lightgray!100}{98.30} & \cellcolor{lightgray!100}{98.60} & \cellcolor{lightgray!100}{98.74} \\
    Ours (128) & 93.92 & 91.81 & 84.68  &  72.26 & 90.03 &  \cellcolor{lightgray!100}{86.54}  &  \cellcolor{lightgray!100}{97.60}  &  94.63 &  96.12  & \cellcolor{lightgray!100}{95.76}  & 94.46  & 94.61 & \cellcolor{lightgray!100}{94.94} \\

    \midrule
    \midrule
    \multicolumn{14}{c}{TPR@0.01FPR} \\
    \midrule
    \tabincell{c}{TreeRing} & 71.92 & 68.93 &  27.97 & 2.95 & 74.51 & 49.26 & 77.73  & 3.44 & 40.59 & 56.51 & 34.50 & 37.22 & 42.74 \\
    \tabincell{c}{S.Signa. (48) } & 1.20  & 0.06 &  26.86 &  0.00 & \underline{99.42} & 25.51 & \underline{99.30} & 0.90  & 50.10 & 0.12 & 71.98 & 37.44 & 36.51 \\
    \tabincell{c}{LaWa (48) } & 85.56 & 84.77 & 78.13 & 2.70 & 91.63 & 68.56 & 85.59 & 76.23 & 80.91 & 66.53 & 79.30 & 80.63 & 75.49 \\
    \tabincell{c}{RoSteALS (56) } & 94.88 & \underline{94.94} & 94.44  &  \underline{86.67} & 70.90 & \underline{88.37}  & 94.64  & 81.80  &  88.22 & \underline{94.54} & \underline{94.78} & 94.72 & \underline{94.68} \\
    \tabincell{c}{S.Stamp  (56) } & \underline{94.91} & 88.60 &  \underline{96.51} &  23.13 & 95.23 & 79.68 & 95.34  & \underline{94.83}  &  \underline{95.09} & 94.40 & 94.72 & \underline{94.78} & 94.63 \\
    \tabincell{c}{DT (56) } & 14.97 & 13.51 & 9.60 &1.53& 0.89 & 8.10 & 12.67 & 9.66 & 11.17 & 13.63 & 11.14 & 11.49 & 12.09 \\
    \midrule
    Ours (48) & \cellcolor{lightgray!100}{100.0} & \cellcolor{lightgray!100}{100.0} & \cellcolor{lightgray!100}{99.94}  &  \cellcolor{lightgray!100}{93.31} & 98.44 & \cellcolor{lightgray!100}{98.34} &  \cellcolor{lightgray!100}{100.0} & \cellcolor{lightgray!100}{100.0}  &  \cellcolor{lightgray!100}{100.0}  & \cellcolor{lightgray!100}{100.0} & \cellcolor{lightgray!100}{99.96} & \cellcolor{lightgray!100}{99.98} & \cellcolor{lightgray!100}{99.98} \\
    Ours (56) & \cellcolor{lightgray!100}{100.0} & \cellcolor{lightgray!100}{100.0} &  \cellcolor{lightgray!100}{100.0} & 85.69  & 99.32 &  \cellcolor{lightgray!100}{97.00}  &   \cellcolor{lightgray!100}{100.0} & \cellcolor{lightgray!100}{100.0}  & \cellcolor{lightgray!100}{100.0}  & \cellcolor{lightgray!100}{100.0} & \cellcolor{lightgray!100}{100.0} & \cellcolor{lightgray!100}{100.0} & \cellcolor{lightgray!100}{100.0} \\
    Ours (64) & \cellcolor{lightgray!100}{100.0} & \cellcolor{lightgray!100}{100.0} & \cellcolor{lightgray!100}{100.0}  & \cellcolor{lightgray!100}{88.31}  & \cellcolor{lightgray!100}{99.94} &  \cellcolor{lightgray!100}{97.65}  & \cellcolor{lightgray!100}{100.0}  &  \cellcolor{lightgray!100}{100.0} & \cellcolor{lightgray!100}{100.0}  & \cellcolor{lightgray!100}{100.0} & \cellcolor{lightgray!100}{99.98} & \cellcolor{lightgray!100}{99.98} & \cellcolor{lightgray!100}{99.99} \\
    Ours (128) & \cellcolor{lightgray!100}{100.0} & \cellcolor{lightgray!100}{100.0} &  \cellcolor{lightgray!100}{99.96} &  83.93 & 98.12 &  \cellcolor{lightgray!100}{96.40}  & \cellcolor{lightgray!100}{100.0}  &  \cellcolor{lightgray!100}{100.0} &  \cellcolor{lightgray!100}{100.0} & \cellcolor{lightgray!100}{100.0} & \cellcolor{lightgray!100}{100.0}  & \cellcolor{lightgray!100}{100.0} & \cellcolor{lightgray!100}{100.0} \\

    \bottomrule
    \end{tabular}
    
    \label{tab: supp attack flickr}
\end{table*}

\section{Examples of LW-Watermarked Images}

\begin{figure}[h]
    \centering  
    \includegraphics[scale=0.11]{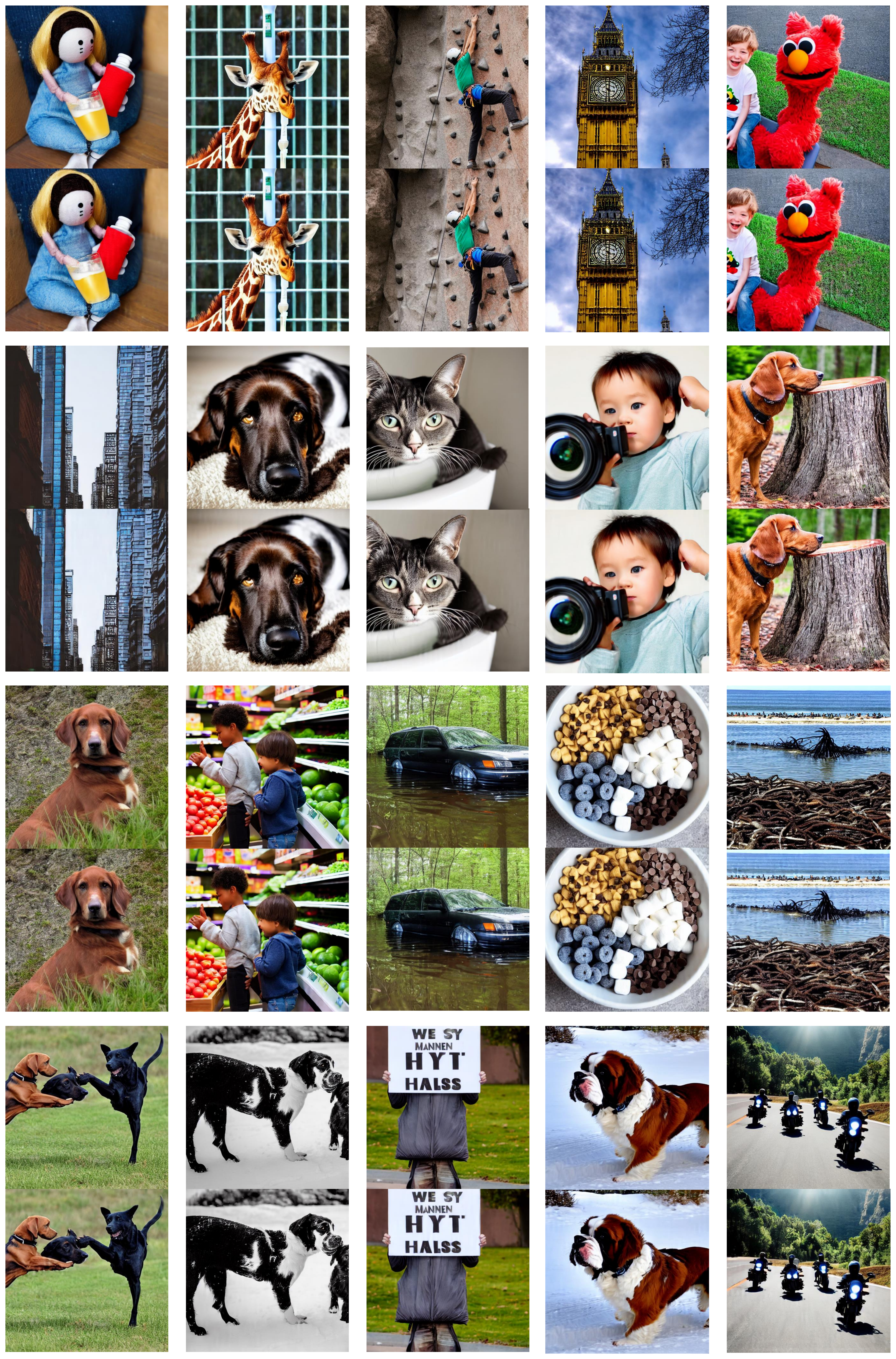}
    \caption{The examples of images watermarked by LW (48 bits). The top of each set is the vanilla image generated by Stable Diffusion, and the bottom is the image watermarked by LW.}  
    \label{fig: supp 48}
\end{figure}

\begin{figure}[h]
    \centering  
    \includegraphics[scale=0.11]{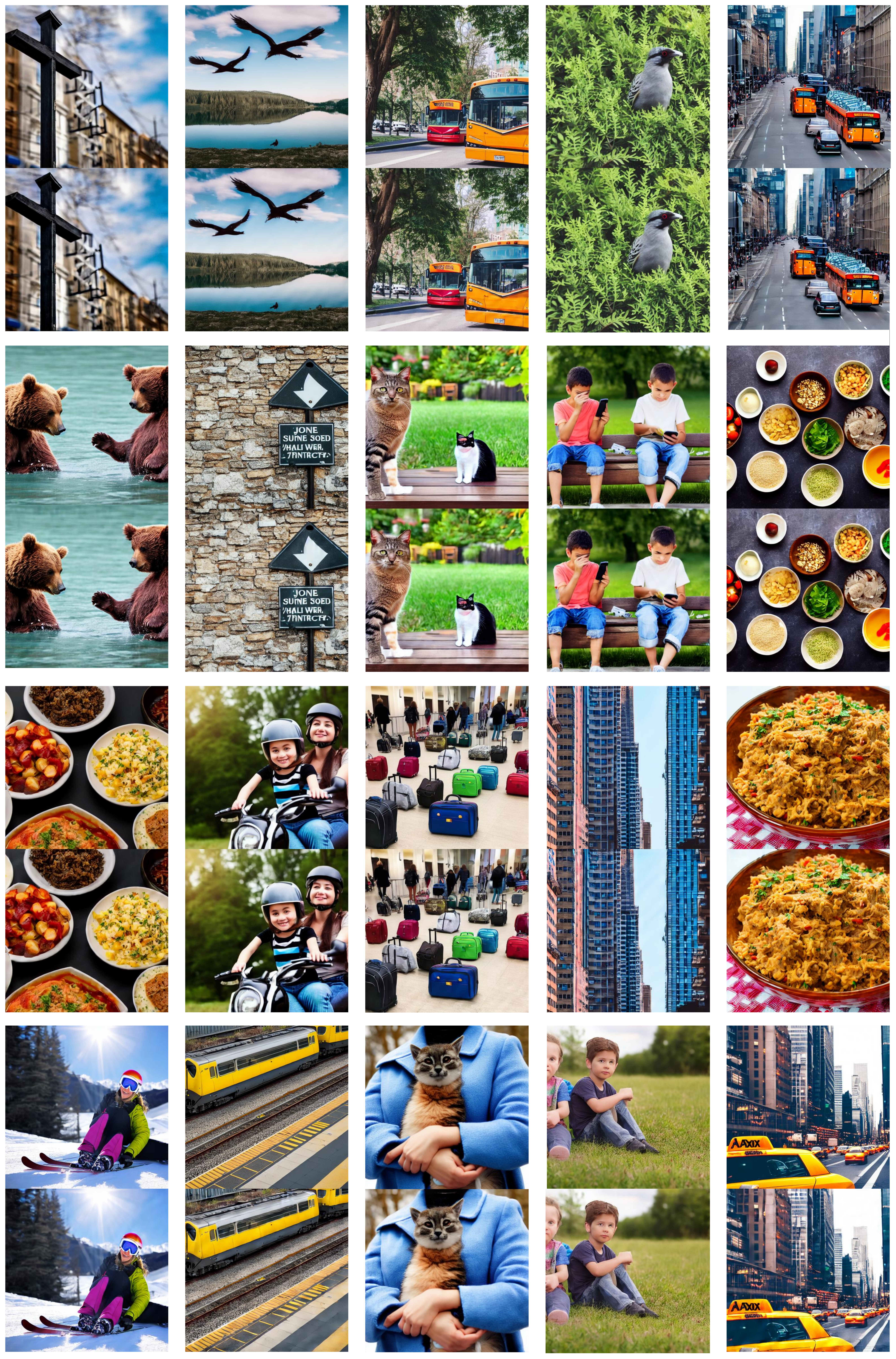}
    \caption{The examples of images watermarked by LW (56 bits). The top of each set is the vanilla image generated by Stable Diffusion, and the bottom is the image watermarked by LW.}  
    \label{fig: supp 56}
\end{figure}

\begin{figure}[h]
    \centering  
    \includegraphics[scale=0.11]{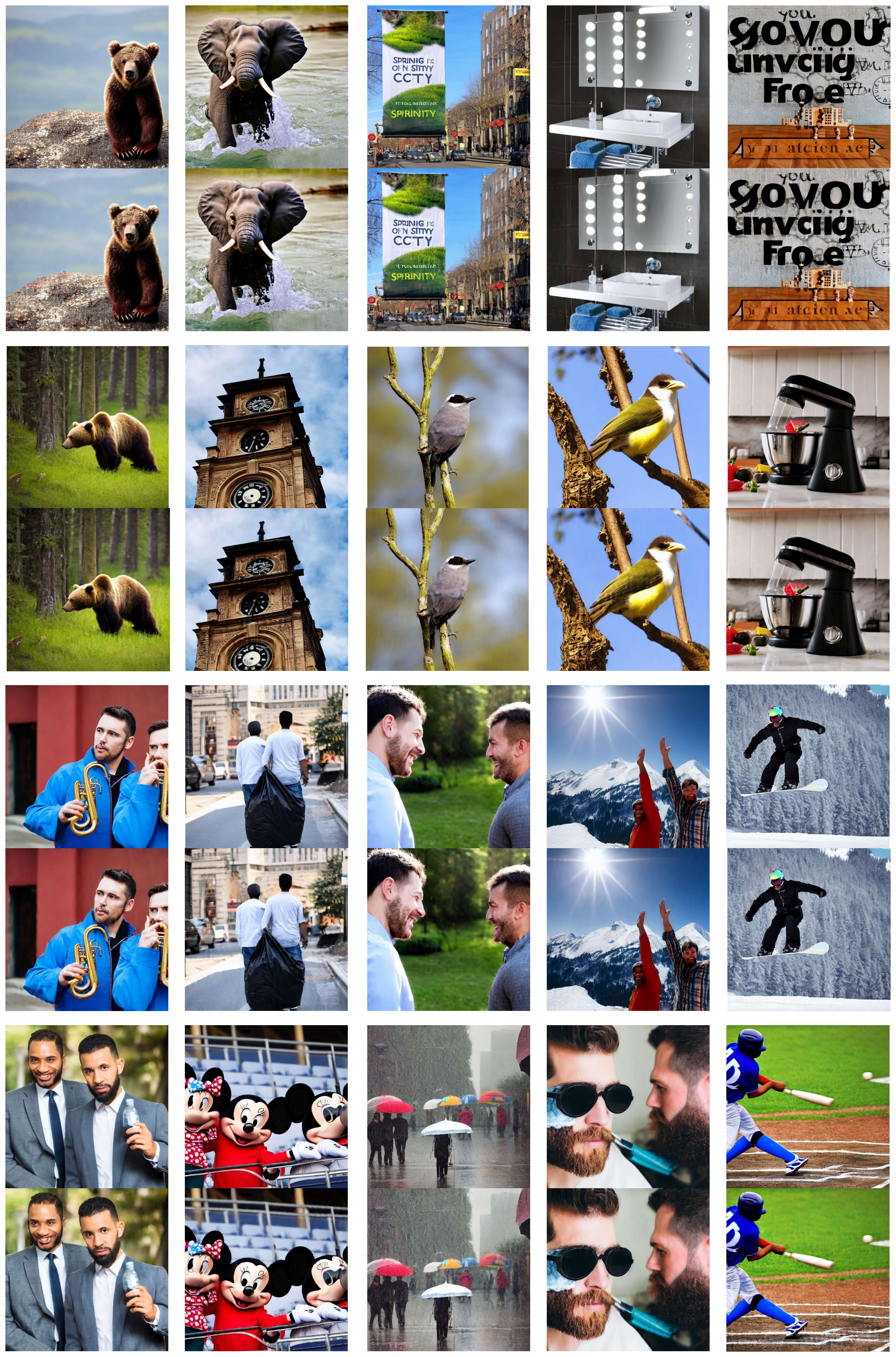}
    \caption{The examples of images watermarked by LW (64 bits). The top of each set is the vanilla image generated by Stable Diffusion, and the bottom is the image watermarked by LW.}  
    \label{fig: supp 64}
\end{figure}

\begin{figure}[h]
    \centering
    \includegraphics[scale=0.11]{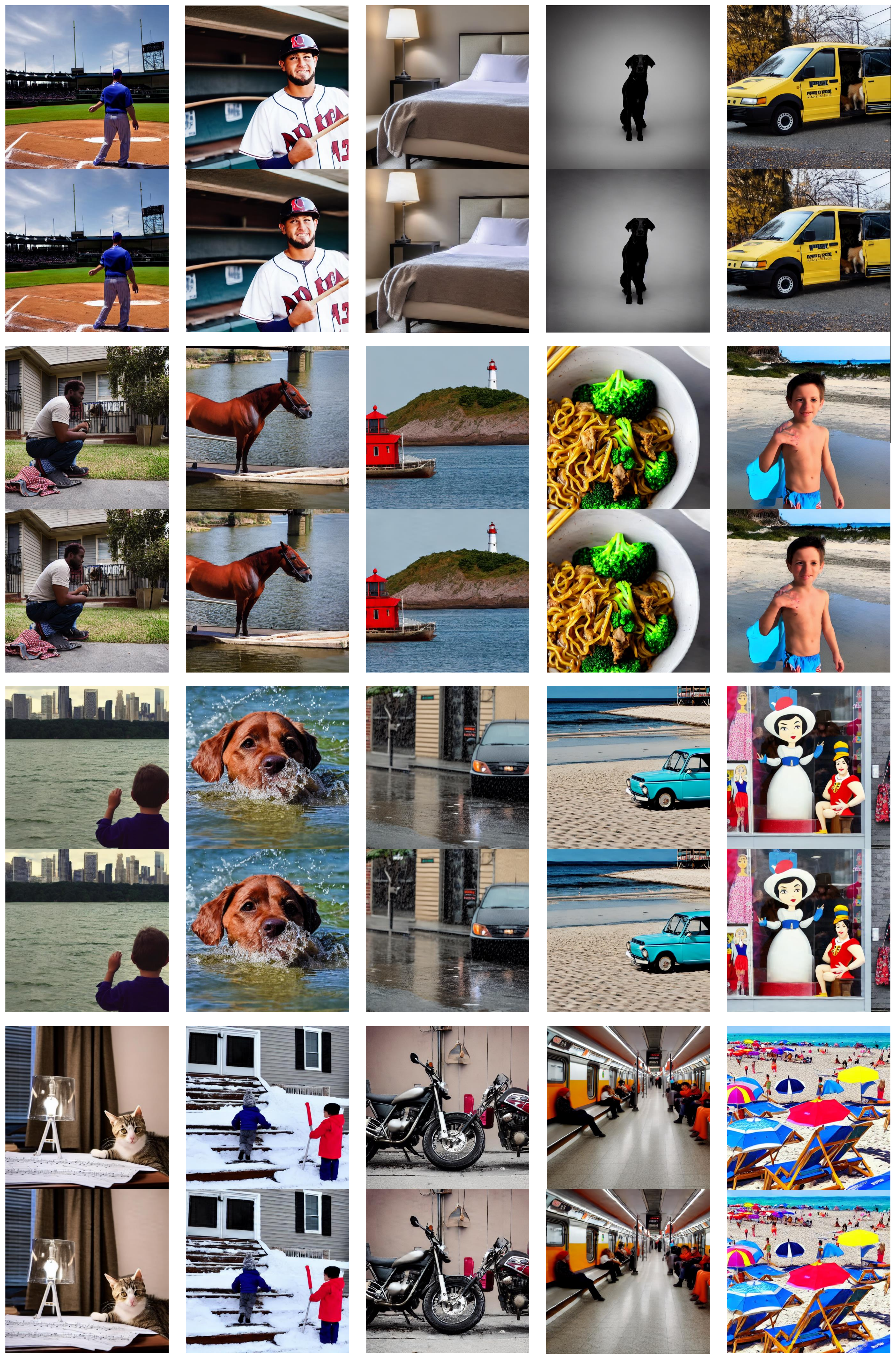}
    \caption{The examples of images watermarked by LW (128 bits). The top of each set is the vanilla image generated by Stable Diffusion, and the bottom is the image watermarked by LW.}  
    \label{fig: supp 128}
\end{figure}

The examples of images watermarked by LW are shown in Fig.\ref{fig: supp 48} (48 bits), Fig.\ref{fig: supp 56} (56 bits), Fig.\ref{fig: supp 64} (64 bits) and Fig.\ref{fig: supp 128} (128 bits).

\end{document}